\documentclass[10pt,twocolumn,letterpaper]{article}

\usepackage[pagenumbers]{cvpr} 

\usepackage{makecell}
\usepackage{multirow}
\definecolor{cvprblue}{rgb}{0.21,0.49,0.74}
\usepackage[pagebackref,breaklinks,colorlinks,allcolors=cvprblue]{hyperref}

\def\paperID{18167} 
\def\confName{CVPR}
\def\confYear{2026}

\author{Rahul Venkatesh\textsuperscript{1,*}, Klemen Kotar\textsuperscript{1,*}, Lilian Naing Chen\textsuperscript{1,*}, Wanhee Lee\textsuperscript{1,*}, \\
Gia Ancone\textsuperscript{1}, Seungwoo Kim\textsuperscript{1}, Luca Thomas Wheeler\textsuperscript{1}, Jared Watrous\textsuperscript{1}, \\ Honglin Chen\textsuperscript{2,\textdagger}, Daniel Bear\textsuperscript{3,\textdagger}, Stefan Stojanov\textsuperscript{4,\textdagger}, Daniel LK Yamins\textsuperscript{1} \\
\fontsize{10.5}{12}\selectfont  \textsuperscript{1} Stanford University, \textsuperscript{2} OpenAI, \textsuperscript{3} Noetik Inc., \textsuperscript{4} Google
}

\newcommand{\lrasseg}{$\textbf{\texttt{PSI}}$\xspace}
\newcommand{\spelkeentity}{$\textbf{\texttt{SpelkeBench}}$\xspace}
\newcommand{\spelkeeditbench}{$\textbf{\texttt{3DEditBench}}$\xspace}

\title{Physical Object Understanding with a Physically Controllable World Model}

\begin{document}

\maketitle
\renewcommand{\thefootnote}{\fnsymbol{footnote}}
\footnotetext[1]{equal contribution}
\footnotetext[2]{work done while at Stanford}
\renewcommand{\thefootnote}{\arabic{footnote}}


\begin{abstract}

A central challenge in visual intelligence is learning the physical structure of scenes from raw videos: how regions form objects and the laws that govern their interactions. Solving these tasks requires world models capable of inferring distributional states of the world from partial observations -- capabilities that current architectures do not provide. We introduce a new class of probabilistic world models that support estimation of the probability of any visual variable, such as appearance and dynamics, conditioned on any other variables. Here, we identify that these models can be trained efficiently with autoregressive sequence modeling, yielding world models from which rich object understanding emerges. First, we demonstrate that our model captures the physical laws governing how objects move by generating multiple plausible future states of the world through sequential inference. Then, by analyzing motion correlations across these futures, we extract objects and articulated object subparts. Having discovered these objects, we show that our world model can manipulate them in 3D. Finally, we demonstrate how physical relationships between objects can be computed from the world model, enabling applications such as Visual Jenga. Our project page and code is available at: \url{https://neuroailab.github.io/psi-website/blog.html}


\end{abstract}


\vspace{-1em}
\section{Introduction}

A central goal in visual intelligence is to build models that understand the physical structure of scenes -- how things move if interacted with, how regions correspond to objects, how objects move in 3D, and the physical relationships among them, such as articulation, support, etc. Achieving this requires probabilistic reasoning that captures how one variable in the scene depends on another -- e.g., how motion at one location influences motion elsewhere, or how forces applied to an object propagate through its structure.

Bringing all these capabilities together within a unified architecture remains a major challenge. Many existing world-modeling approaches operate on prompts such as text~\cite{touvron2023llamaopenefficientfoundation}, actions~\cite{bruce2024genie}, or global scene embeddings~\cite{rombach2022high, agarwal2025cosmos} -- which do not provide a way to isolate or query how local scene variables (e.g., appearance, depth, motion) influence one another. Understanding these relationships is essential for reasoning about objects and their physical interactions.

Here, we solve this problem by introducing a new class of world models, where we represent the state of the world with local variables and predict the probability distribution of each such variable, given any other variables. We show that this model concept can be formulated as a GPT-style~\cite{radford2018improving} next token prediction sequence model, allowing efficient and scalable model training.  Under this framework, we build a Physically controllable World Model, with tokens describing visual scenes such as RGB tokens that encode appearance, flow tokens that encode dynamics, and camera tokens that encode viewpoint changes between frames. 

Our world model supports a wide range of inference pathways by treating different scene variables as probabilistic structures that can either be inferred from observations or used to condition future predictions. For example, optical flow can be generated as an intermediate representation to model plausible scene dynamics conditioned on sparse pokes, or supplied as a conditioning signal to render future appearance states of the world. More generally, our framework can be used to integrate multiple structured representations of a scene—such as appearance, motion, geometry, and actions—within a unified probabilistic framework. We therefore refer to this framework as a Probabilistic Structure Integrator (\lrasseg), emphasizing its ability to represent, predict, and compose diverse scene structures through probabilistic inference.


We then use \lrasseg to extract a wide range of physical object understanding. First, its predicted distributions provide motion statistics, highlighting which regions are likely to move and by how much. Next, sequential inference over flow and RGB tokens produces multiple plausible futures, capturing uncertainty and dynamics in the visual world. From these imagined futures, we derive object-like entities by using motion-correlation analysis to group pixels that move together, achieving state-of-the-art object segmentation scores on SpelkeBench~\cite{SpelkeBench}. Further, we show how articulated subparts can be found, outperforming prior methods on DragAMove~\cite{baumann2025if}. These discovered objects can then be consistently manipulated in 3D, achieving state-of-the-art results on 3DEditBench~\cite{lee20253d}. Finally, we demonstrate how the world model reasons about physical relationships between objects to perform tasks like Visual Jenga~\cite{bhattad2025visual}.  

\section{Related Works}

\textbf{Visual world models:} Recent world models provide strong generative and perceptual abilities -- they excel at producing long-form video~\cite{bruce2024genie,agarwal2025cosmos}, performing instruction-conditioned generations~\cite{lin2024video,agarwal2025cosmos}, and offering rich multimodal reasoning~\cite{team2025gemma}. Yet these systems fall short of the flexible conditional inference that deeper physical scene understanding requires~\cite{wu2024physical}. They do not expose a mechanism for physically grounded queries -- such as how local scene variables change under specific physical interventions. \lrasseg overcomes this by modeling the visual world as a probabilistic model over local variables, enabling fine-grained causal reasoning for a range of scene-understanding tasks. Further, while classical world models are typically defined as action-conditioned predictors~\cite{ha2018world}, \lrasseg is best viewed as a ``poor man's world model"~\cite{kotar2025worldmodelingprobabilisticstructure}, where expensive-to-obtain true action
data is proxied by cheap visual data patches like optical flow and camera conditioning that encode approximations to agent actions.

\textbf{Specialized object understanding models:} Several prior works target aspects of object and scene understanding, typically in specialized settings. For object discovery, supervised segmentation models such as SAM2~\cite{ravi2024sam2} rely on costly annotations, but capture appearance-based groupings rather than physically coherent entities~\cite{zhang2023understanding, ji2304segment}. Self-supervised  methods like CutLER~\cite{wang2023cut} and ProMerge~\cite{li2024promerge} combine attention maps from pretrained encoders~\cite{oquab2023dinov2} to identify object regions, but remain brittle in complex natural scenes. Object editing systems -- including drag-based manipulation~\cite{wu2024draganything, gillman2025force, chen2025perception, baumann2025if} and depth-conditioned diffusion models~\cite{pandey2024diffusion, gu2025diffusionshader3dawarevideo} -- understand how objects move in 3D, but their performance degrades in cluttered scenes~\cite{chen2025perception}. Methods like Visual Jenga~\cite{bhattad2025visual} infer physical relationships between objects using a combination of off-the-shelf models. In contrast, \lrasseg is a unified, physically controllable world model that discovers objects, understands how they move in 3D, and infers physical relationships in a zero-shot way without requiring specialized architectures.


\section{\lrasseg architecture and training details}
\label{sec:architecture_details}

In this section, we first introduce the formulation of world models as probabilistic graphical models~\cite{kotar2025worldmodelingprobabilisticstructure}. We then describe how these models can be realized using a sequence prediction architecture. Finally, we present a physically controllable visual world model built under this framework.

\begin{figure*}[!t]
  \centering
  \includegraphics[width=0.98\textwidth]{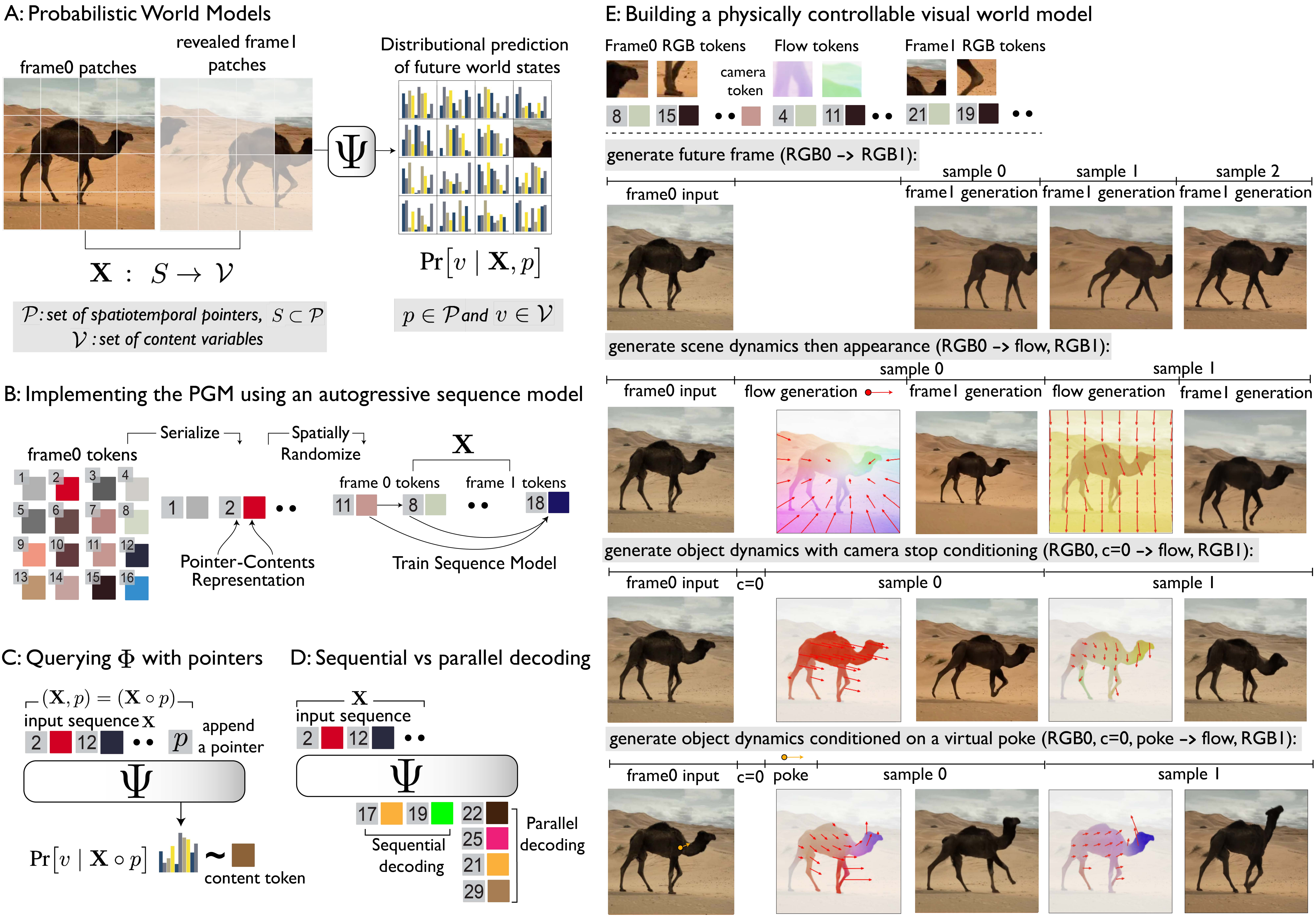}
  \captionsetup{labelfont=bf}
  \vspace{-0.8em}
  \caption{\textbf{\lrasseg Architecture.} Here, \textbf{Figure A} illustrates a probabilistic model of the visual world, which predicts distributions over visual variables such as appearance, motion, and depth. \textbf{Figure B} shows how this can be implemented by training a sequence model on a pointer–content representation of visual data. In \textbf{Figure C}, we show how the learnt sequence model can be queried to obtain distributional estimates over visual variables at specified locations. Building on this, \textbf{Figure D} shows how the model can also produce visual variables, either in parallel across the scene or sequentially. Finally, in \textbf{Figure E} we describe \lrasseg, a physically controllable visual world model built under this framework, and its various inference pathways using RGB, optical flow and camera motion tokens. }
  \label{fig:main}
  \vspace{-1em}
\end{figure*}

\textbf{Probabilistic World Models.}  Building models of the visual world involves learning distributions of future states of the world, conditioned on partial observations. To this end, we construct a probabilistic world model whose basic concept is to (1) partition the state of the world into a set of local variables and then (2) learn to predict the conditional probability distribution of each such variable, conditioned on any subset of these variables. Formally, we start with a data format that is described by a finite set of
spatiotemporal pointer locations $\mathcal{P}$ and a finite vocabulary $\mathcal{V}$ of content values. A partial description of the world is specified as a datum $\mathbf{X}: S \rightarrow \mathcal{V}$ that maps a subset of spatiotemporal indices $S \subset \mathcal{P}$ to content variables. The learned model $\Psi$ accepts this datum, and an unobserved query location $p \notin \mathrm{dom}(\mathbf{X})$ and returns a probability distribution $\text{Pr}[v \,|\, \mathbf{X},p]$, capturing the model’s belief over what content could appear at $p$, given what is observed (see Figure~\ref{fig:main}a):
\begin{equation}
\vspace{-0.3em}
\Psi: (\mathbf{X},\, p \notin \mathrm{dom}(\mathbf{X})) \longmapsto 
\big\{\text{Pr}\big[v\mid \mathbf{X}, p\big] \;:\; v \in \mathcal{V}\big\}.
\label{eq:pgm-conditional}
\end{equation}
Conceptually, this formulation is reminiscent of Probabilistic Graphical Models (PGMs), systems in which the distribution of any variable can be computed conditioned on any subset of other variables in the system~\cite{koller2009probabilistic}.
 
\textbf{Implementing $\Psi$ with Autoregressive Sequence Prediction.} Because it has historically been difficult to learn PGM models~\cite{frey2013learning} -- one of the reasons they are not popular in modern deep learning -- a key question we face is: how exactly can we formulate a neural-network learning objective to efficiently learn $\Psi$. Our core insight is that the problem of learning arbitrary inter-variable conditional relationships can be reformulated as a GPT style next token prediction sequence model.  Specifically, we serialize the datum, $\mathbf{X}$, as a sequence of interleaved pointer and content tokens, $[p_0, v_0, ...p_k, v_k]$ (see Figure~\ref{fig:main}b). The PGM formulation in equation~\ref{eq:pgm-conditional} reduces to the standard autoregressive framework, where probability distributions over content tokens are obtained by appending a pointer at the end of the sequence (see Figure~\ref{fig:main}c): 
\begin{equation}
\vspace{-0.3em}
\Psi(\mathbf{X}, p)
= \Psi(\mathbf{X} \circ p)
\equiv 
\text{Pr}\left[
  v \,\middle|\,
  \mathbf{X} \circ p
\right]
\label{eq:ar-bridge}
\end{equation}
This equivalence of PGMs with sequence models allows us to leverage standard architectures that have proven highly effective in building powerful models in AI, in particular language modeling.

The use of pointer tokens enables sequences to be constructed in arbitrary spatiotemporal order. Each next-token prediction, conditioned on previously observed tokens in the sequence, corresponds to estimating a variable given a partial state -- sampling the underlying PGM. Training across various orderings amortizes this process, allowing the model to learn the full PGM, i.e., the joint distribution over all variables.

\textbf{Building a physically controllable visual world model.} The model described above is a generic construction that can in principle operate on variables from any modality. In this work, we use it to build a physically controllable visual world model, \lrasseg (Figure~\ref{fig:main}e). We do this by constructing a content vocabulary
$\mathcal{V}$ with tokens describing visual scenes: RGB tokens ($\mathcal{V}^{(\text{rgb})} = \{r_k...\}$) that encode appearance, flow tokens ($\mathcal{V}^{(\text{flow})} = \{f_k...\}$) that encode dynamics, and camera tokens ($\mathcal{C} = \{c_k...\}$) that encode viewpoint changes between frames. The RGB and flow tokens are obtained by training shallow convolutional quantizers on spatial patch data (see Supplement Section 3 for details of quantizer construction), the camera tokens are obtained by binning 6DOF transforms. Additionally, the set of spatiotemporal pointer tokens $\mathcal{P}$ is partitioned into two sets ($\mathcal{P}^{(\text{rgb})} = \{p_k^{\text{(rgb)}}..\}$ and $\mathcal{P}^{(\text{flow})} = \{p_k^{\text{(flow)}}...\}$), for the RGB and flow modalities, respectively. Using these tokens, we construct training sequences of the form: 
\begin{equation}
\vspace{-0.3em}
\mathbf{X} = \mathbf{r}^0\, \circ \,[c] \, \circ \mathbf{f} \,\circ \mathbf{r}^1
\end{equation}
where, $\mathbf{r}^0 = [p_k^{\text{(rgb)}}, r_k, ..]$ and $\mathbf{r}^1=[p_k^{\text{(rgb)}}, r_k, ..]$ are RGB pointer-content sequences of two frames of a video, $\mathbf{f}=[p_k^{\text{(flow)}}, f_k, ..]$ is a sequence of flow pointer-content tokens, $c$ is a camera motion token, and $\circ$ denotes concatenation. 
During training, $\mathbf{r}^0$, $\mathbf{r}^1$ and $\mathbf{f}$ are constructed in arbitrary spatial order, camera tokens are provided when available, and flow tokens are randomly masked with a mask ratio between 0-1. This allows the model to learn to predict $\mathbf{r}^1$ given $\mathbf{r}^0$ or conditioned on any amount of flow tokens.

\begin{figure*}
  \centering
  \includegraphics[width=\linewidth]{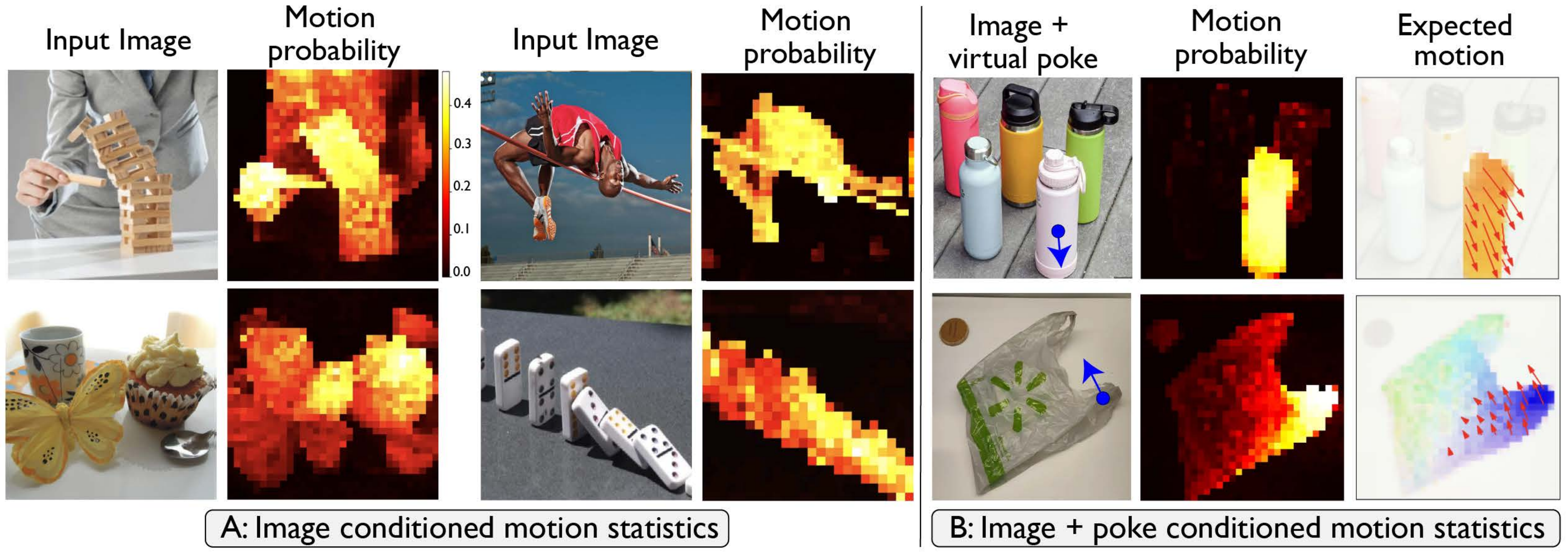}
  \vspace{-2em}
  \caption{
  \textbf{Motion statistics computed in parallel.} In \textbf{Figure A}, we show motion probability computed given the input image and camera stop conditioning. They clearly highlight the parts of the scene that are likely to move. In \textbf{Figure B}, we show probability of motion and expected motion maps for a scene with rigid object (top), and a scene with deformable object (bottom). While the rigid object shows uniform responses across the extent of the object, the deformable object correctly shows localized responses near the poke point. } 
  \label{fig:p_motion_exp}
  \vspace{-1.4em}
\end{figure*}

\textit{Training details.} 
We implement $\Psi$ as a 7B-parameter GPT transformer, and train with next-token prediction cross entropy loss, supervising only the content token (and not the pointer token) sequence elements. The training dataset consisted of 3 million real-world RGB video clips, yielding a total of approximately 1.4 trillion tokens.
Training used a batch size of 512 for 1.5M steps using a Warmup-Stable-Decay schedule~\cite{wen2024understanding}. More details on the model architecture and training process are in Supplement Section 3.

\textbf{Inference pathways.} As trained, $\Psi$ supports a wide variety of inference pathways. For example, as shown in Figure~\ref{fig:main}a, a few spatiotemporal patches of the camel are revealed in a future frame and $\Psi$ predicts a spatial map of distributions over scene variables at other unrevealed locations. Sampling from these yield multiple plausible completions for the rest of the camel.

More generally, at inference we roll out by iteratively choosing undecoded pointers and sampling their content. The type of pointer token we choose (i.e. from $\mathcal{V}^{\text{rgb}}$ or $\mathcal{V}^{\text{flow}}$) decides the modality we decode. For instance, given $\mathbf{r}^0$, if we want to decode a flow token, we would append a pointer $p_k^{\text{(flow)}}$ to the end of the sequence: $\hat{f} \sim \Psi(\mathbf{r}^0 \circ  p_k^{\text{(flow)}})$. Repeating this sequentially gives us a complete set of pointer-flow tokens, $\hat{\mathbf{f}} \overset{\text{seq}}{\sim} \Psi(\mathbf{r}^0; \text{flow})$, $\overset{\text{seq}}{\sim}$ indicates \textit{sequential} sampling, and the flow argument provided to $\Psi$ indicates that we are decoding flow tokens. These flow tokens can then be decoded into pixels using a quantizer decoder. A similar procedure can be followed for RGB tokens.  






Interleaving flow tokens between the RGB tokens of two frames enables the use of flow either as a prediction target (i.e. generating plausible scene dynamics) or as a conditioning signal (i.e. rendering out future appearance states of the world, conditioned on dynamics). The camera motion token allows us to specify changes in viewpoint. In Figure~\ref{fig:main}e we show various inference pathways of our model:
\begin{itemize}
    \item  $\hat{\mathbf{r}} \overset{\text{seq}}\sim \Psi(\mathbf{r}^0;\text{rgb})$: generating future scene appearance.
    \item  $\hat{\mathbf{f}} \overset{\text{seq}}\sim \Psi(\mathbf{r}^0;\text{flow})$: generating future scene dynamics.
    \item  $\hat{\mathbf{f}} \overset{\text{seq}}\sim \Psi(\mathbf{r}^0 \circ [c=0],\text{flow})$: generating plausible future object dynamics, with camera stop conditioning.
    \item $\hat{\mathbf{f}} \overset{\text{seq}}\sim \Psi(\mathbf{r}^0 \circ [c=0] \circ \mathbf{f}^{\text{sparse}};\text{flow})$: generating a plausible flow completion, given a sparse poke \(\mathbf{f}_{\text{sparse}} = [(p, f^{sp})] \).
    \item $\hat{\mathbf{r}^1} \overset{\text{seq}}\sim \Psi(\mathbf{r}^0 \circ [c=0] \circ \hat{\mathbf{f}}; \text{rgb})$: generating appearance changes resulting from object dynamics, $\mathbf{f}$.
\end{itemize}


As our model predicts a distribution over visual variables, we can sample multiple plausible generations for each inference pathway. In the next section, we describe how these inference pathways enable a wide range of scene understanding applications. 


\textbf{Parallel generation.}  
The procedure we described above uses the standard sequential sampling used in sequence models like LLMs. However, in a variety of applications it can be advantageous to run $\Psi$ in \emph{parallel} mode, where inference on undecoded locations is performed by independently evaluating equation~\ref{eq:ar-bridge} for multiple pointers $p$ simultaneously.  Sequential sampling maximizes quality, as each prediction conditions on all previously generated patches, capturing causal dependencies in complex settings where conditional independence between patches cannot be assumed (e.g., articulated objects such as horses). Parallel generation assumes conditional independence between undecoded locations, trading off efficiency for quality. We illustrate this procedure in Figure~\ref{fig:main}d.





\begin{figure*}
  \centering
  \includegraphics[width=\linewidth]{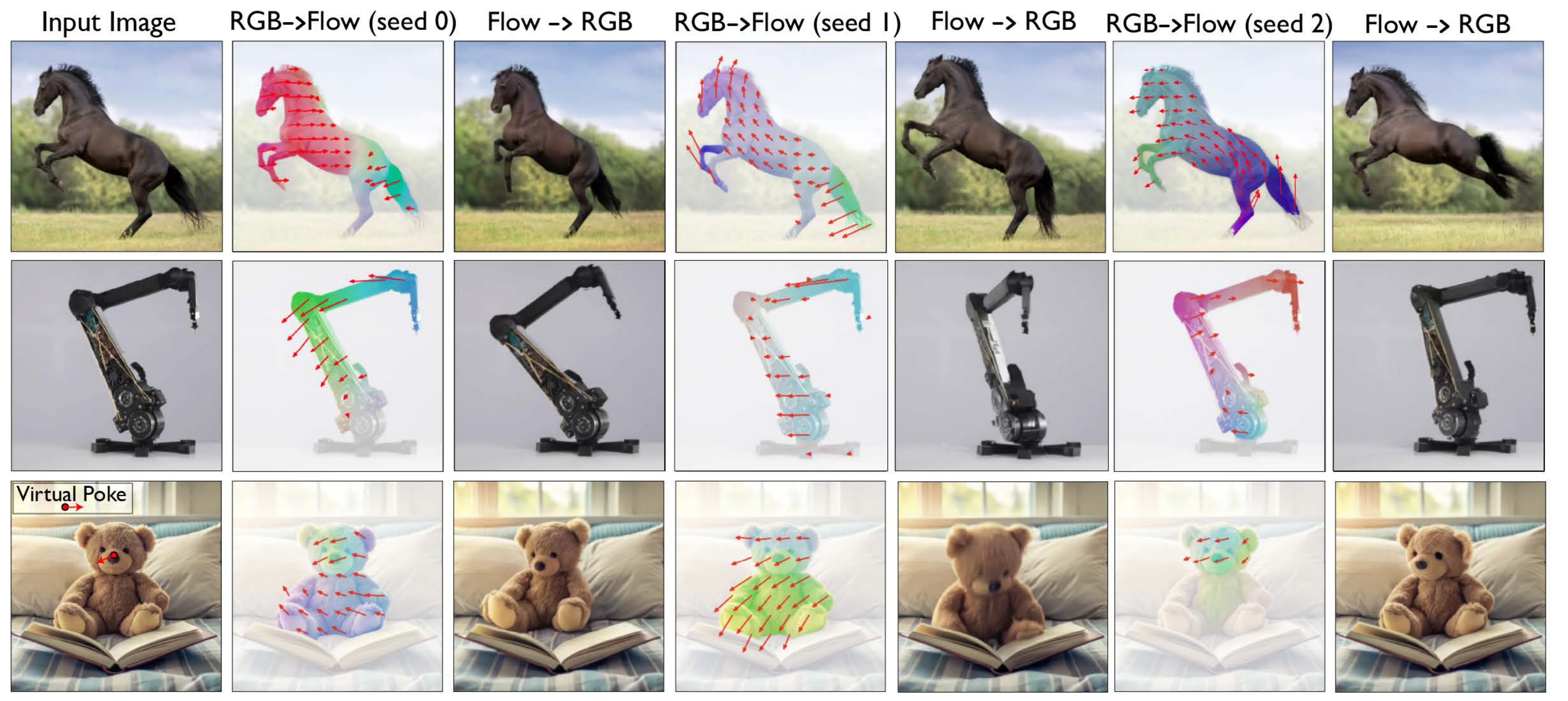}
  \vspace{-2em}
  \caption{
  \textbf{Sequential generation of plausible object dynamics and appearance.} We show that our model can generate multiple physically plausible scene motions and render them into future appearance states -- capturing the true dynamics of the physical world for complex objects. In rows 1–2, the model infers plausible motion patterns directly from a single input image. In row 3, specifying a motion for a part of the object (such as the head of the bear) generates diverse, yet physically consistent responses for the rest of the body.} 
  \label{fig:uncond_gen}
  \vspace{-1.0em}
\end{figure*}


\section{Applications}

In this section, we present a wide range of object-understanding applications enabled by our world model. We begin by using the model’s predicted distributions to estimate motion statistics -- highlighting which regions are likely to move and by how much (Section~\ref{sec:motion_prob}). We then show multiple plausible future motions of an input scene, generated through sequential inference (Section~\ref{sec:seq_gen}). From these imagined futures, we compute motion-correlation statistics to extract object-like entities as groupings of pixels that move together (Sections~\ref{sec:point_prompted}, ~\ref{sec:unprompted} and ~\ref{sec:articulated}). We then demonstrate how these objects can be manipulated in 3D (Section~\ref{sec:manip_3d}). Additionally, we describe a procedure to uncover physical relationships across various objects in the scene (Section~\ref{sec:support}), enabling applications like Visual Jenga~\cite{bhattad2025visual}. Finally, we analyze the statistical significance of our results in Section 7 of the Supplement.

\subsection{Parallel estimation of motion statistics}
\label{sec:motion_prob}

\textit{Motion probability maps.} For many practical applications, particularly in robotics, it is valuable to know which regions of the scene respond to physical interaction (e.g. movable objects) and those that do not (e.g. floors and walls). Our world model can be queried to produce an estimate of the \emph{probability of motion} at a pointer $p$, by summing probabilities, $\text{Pr}(f_j \,|\, \mathbf{X} \circ p) = \Psi (\mathbf{X} \circ p)$ over the token set $\mathcal{F}_{\text{motion}}$, consisting of flow tokens corresponding to flows greater than a certain magnitude:
\begin{equation}
\label{eqn:p_motion}
\vspace{-0.4em}
\mathbb{P}_{\text{motion}}[p] = \sum_{f_j \in \mathcal{F}_{\text{motion}}} \text{Pr}(f_j \,|\, \mathbf{X} \circ p).
\vspace{-0.4em}
\end{equation}
$\mathbb{P}_{\text{motion}}$ can be computed in parallel across all pointers, forming a 2D heatmap of regions likely to move. We append a zero camera motion token, ensuring that the predicted dynamics reflect only the effects of physical interaction. Given the input image, the map reflects which regions would move if interacted with (Figure~\ref{fig:p_motion_exp}a) and when conditioned on a virtual poke, the resulting motion map highlights the regions that would respond to that intervention (Figure~\ref{fig:p_motion_exp}b).   


\textit{Expected motion maps}. While $\mathbb{P}_{\text{motion}}$ tell us which parts of the scene will move, often we need to reason about the result of virtual interactions (i.e., where each point moves). To do so, we compute the expected motion at a location, $p$ as the probability-weighted average of flow vectors \( \mathbf{v}_j \), where each \( \mathbf{v}_j \) maps to token \( f_j \):
\begin{equation}
\vspace{-0.7em}
\mathbb{E}_{\textrm{motion}}[p] = \sum_{f_j \in \mathcal{V}^{(flow)} } \text{Pr}(f_j \,|\, \mathbf{X} \circ p) \cdot \mathbf{v}_j \quad 
\end{equation}
$\mathbb{E}_{\textrm{motion}}$ is computed in parallel across all pointers, and can be obtained for any number of virtual pokes (See Figure~\ref{fig:p_motion_exp}b).


\subsection{Sequential generation of plausible future motions of a scene}
\label{sec:seq_gen}

While the parallel motion statistics described above provide useful estimates of how spatial locations are expected to move, they implicitly assume conditional independence across locations. To capture the rich dependencies present in complex, multi-part objects, \lrasseg supports sequential generation, as discussed in Section~\ref{sec:architecture_details}. By sampling tokens one at a time, the model produces diverse motion fields that naturally respect spatial dependencies. We illustrate this capability on multiple challenging objects in Figure~\ref{fig:uncond_gen}, on both poke-conditioned and unconditional generation.  



\subsection{Point-prompted movable object segmentation using motion correlation statistics} 
\label{sec:point_prompted}

\begin{figure*}[!t]
  \centering
  
  \begin{subfigure}[t]{\textwidth}
  \centering
  \includegraphics[width=0.87\linewidth]{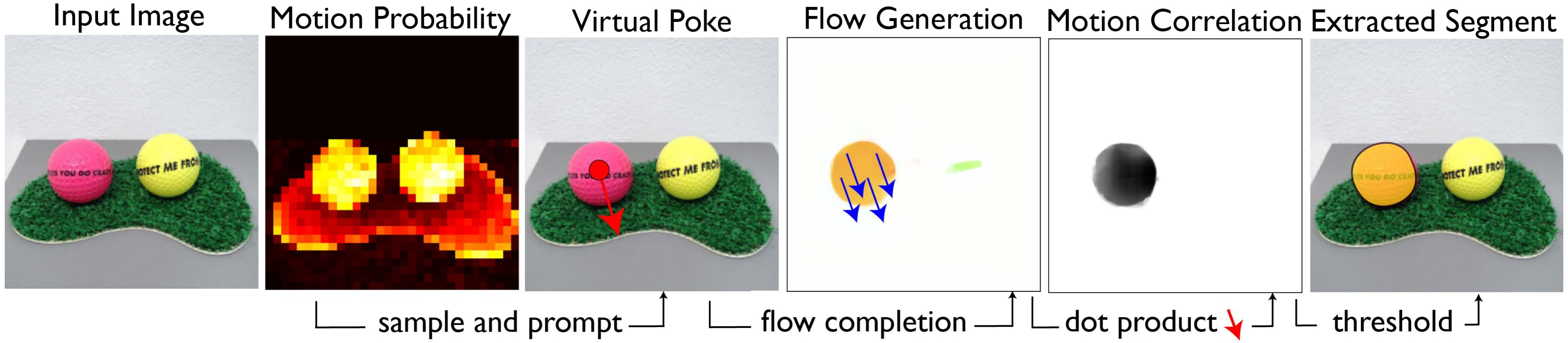}
  \caption{Movable object discovery using correlated motion statistics in poke conditioned sequential flow generations.}
  \label{fig:spelke_discovery}
  
    \vspace{0.3em}
    
  \end{subfigure}
  \begin{subfigure}[t]{\textwidth}
    \centering
    \includegraphics[width=0.92\linewidth]{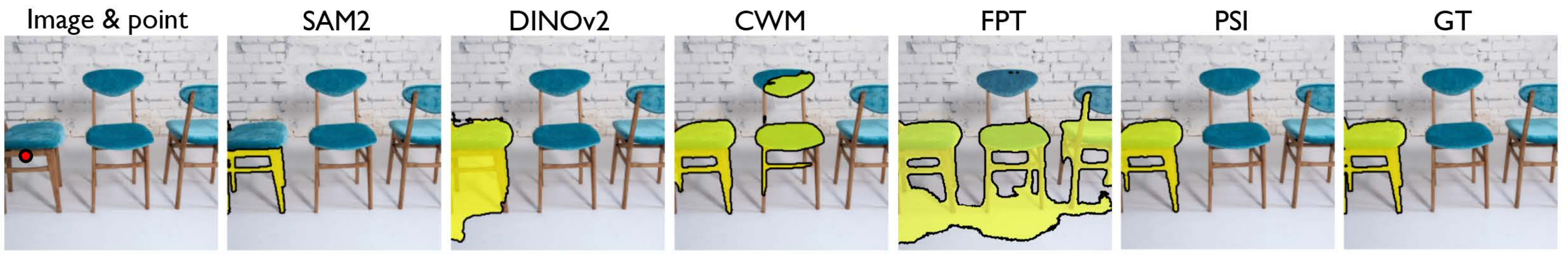}
    \caption{Point-prompted movable object segmentation.}
    \label{fig:pointsegres}
  \end{subfigure}

  \vspace{0.5em}

  \begin{subfigure}[t]{\textwidth}
    \centering
    \includegraphics[width=0.92\linewidth]{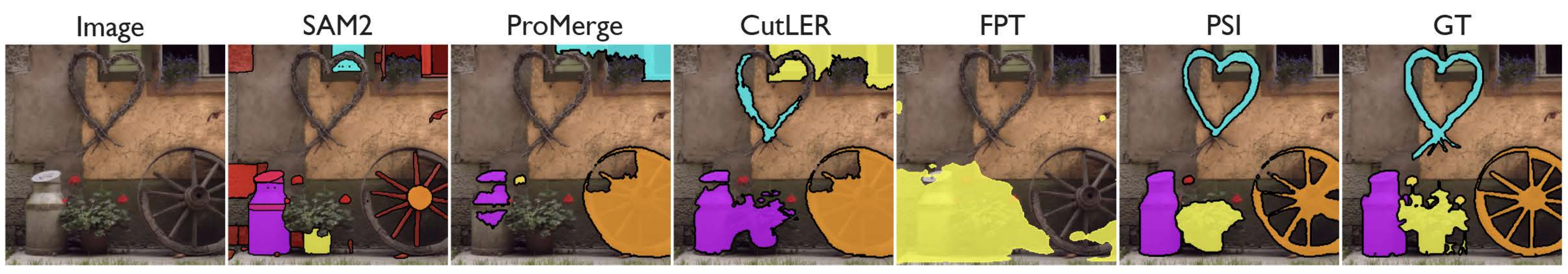}
    \caption{Unprompted movable object segmentation.}
    \label{fig:autosegqual}
  \end{subfigure}

  \vspace{0.5em}

  \begin{subfigure}[t]{\textwidth}
    \centering
    \includegraphics[width=0.92\linewidth]{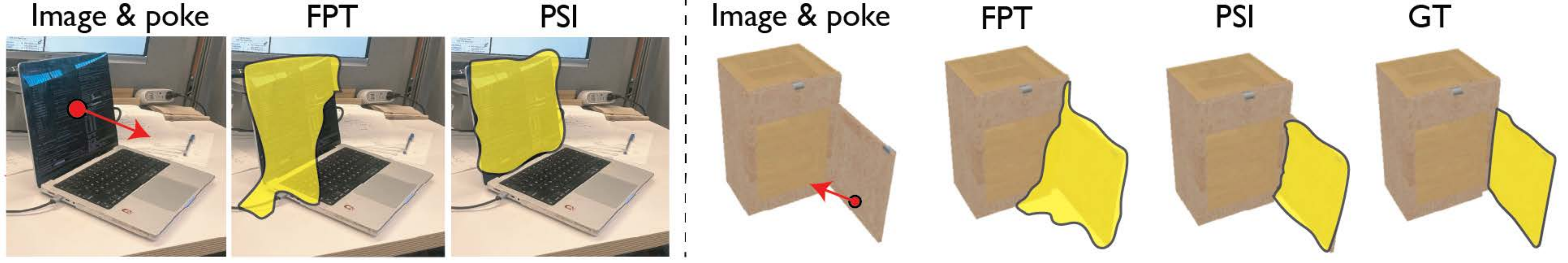}
    \caption{Poke-prompted articulated part segmentation.}
    \label{fig:dragamove}
  \end{subfigure}
   \vspace{-0.7em}
  \caption{\textbf{Movable object segmentation tasks with \lrasseg.} In this figure, we show qualitative results on the segmentation tasks described in Sections~\ref{sec:point_prompted},~\ref{sec:unprompted} and~\ref{sec:articulated}. \lrasseg  obtains segments that align better with the notion of what moves together in the physical world.}
  \label{fig:segqual_all}
  \vspace{-0.5em}
\end{figure*}

The diverse motion samples produced through sequential generation allows us to probe the scene’s causal structure by analyzing motion-correlation statistics. We use this to identify \textit{movable objects} -- coherent groupings of pixels that move in unison. To extract these given a point prompt \(p \), we compute the \textit{motion correlation map}, as the dot product between a poke vector $\mathbf{v}_j$ that maps to a virtual poke $f_j$, and the flow generation resulting from the poke: \(\hat{\mathbf{f}}_j \overset{\text{seq}}\sim \Psi( \mathbf{r}^0 \circ [c=0] \circ [p, f_j], \text{flow})\) -- camera stop conditioning is used ($c=0$) to discount camera motion. Computing the expected motion correlation, over $N$ pokes, and thresholding it yields a movable object (see Figure~\ref{fig:spelke_discovery}): 
\begin{equation}
\vspace{-0.5em}
\bar{\text{dot}} =  \frac{1}{N} \sum_{j=1}^N  \mathbf{v}_j \cdot  \hat{\mathbf{f}}_j 
\label{eqn:dot_bar}
\end{equation}
\textit{Results.} We use the SpelkeBench dataset~\cite{SpelkeBench} -- a benchmark with movable object annotations for evaluation. We test our model with $N=8$ pokes and measure segment boundary precision using mean intersection over union (mIoU). We also report Average Recall (AR) which measures detection accuracy as the fraction of segments with $\text{IoU} \geq \tau \in [0.5, 0.99]$. \lrasseg obtains state-of-the-art results on SpelkeBench, as shown in Table~\ref{tab:combined_pointseg}. Qualitative results in Figure~\ref{fig:pointsegres} indicate that self-supervised methods leveraging DINO’s~\cite{oquab2023dinov2} features such as CutLER~\cite{wang2023cut} and Promerge~\cite{li2024promerge} tend to merge multiple instances of the same category and SAM2~\cite{ravi2024sam2} often segments object subparts based on texture that do not move independently. We test other world models such as CWM~\cite{venkatesh2024understanding}, which performs physical interactions through patch motion counterfactuals, and methods like Force Prompting~\cite{gillman2025force}, Flow poke transformer (FPT)~\cite{baumann2025if} and Perception-as-Control (PasC)~\cite{chen2025perception}, which do so with 2D drag vectors. While PasC generalizes poorly to the complex scenes in SpelkeBench, CWM and FPT perform reasonably well, emerging as the strongest self-supervised baselines. Models such as EISEN~\cite{chen2022unsupervised}, which learn segments from motion, and Adaptive Slot Attention~\cite{fan2024adaptive}, do not generalize beyond the simple datasets they were trained on. \lrasseg’s segments better align with the notion of objects as units that move together. See Supplement Section 1.5 for more qualitative results. 




\begin{figure*}[htbp!]  
  \centering
  \includegraphics[width=0.94\linewidth]{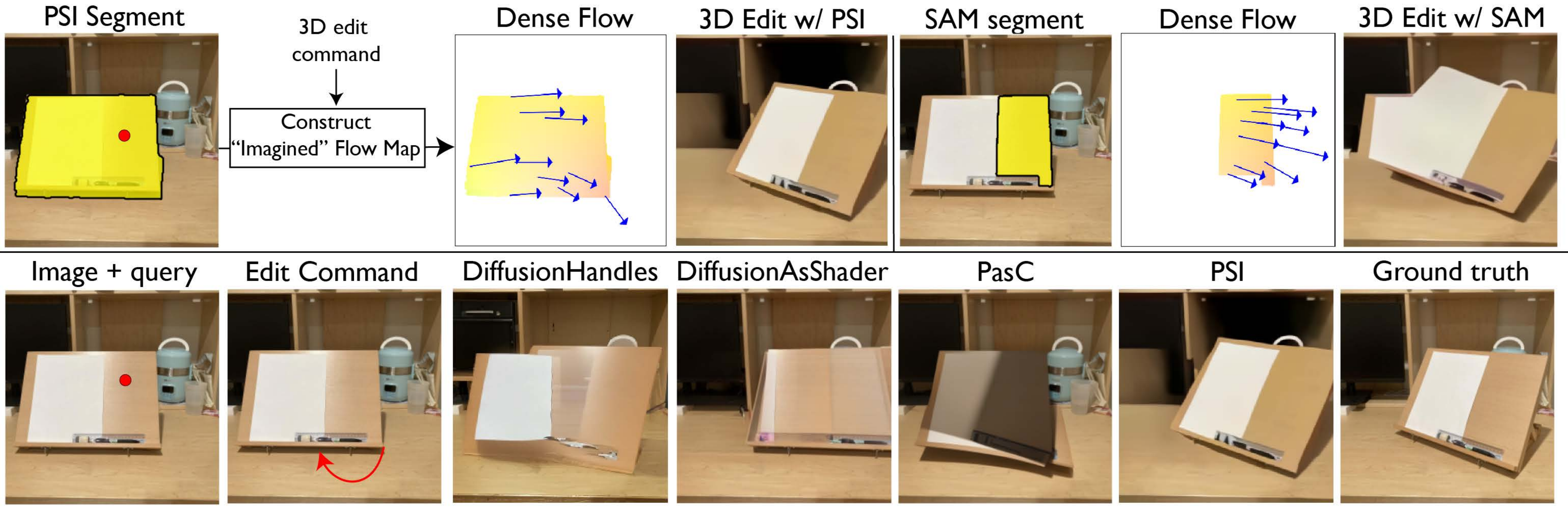}
  \vspace{-0.5em}
  \caption{\textbf{Manipulating discovered objects in 3D}.  On the \textbf{top} we show the pipeline of object manipulation: we extract a segment with a point prompt, and given a 3D Edit, we produce a dense flow map reflecting the transformation, and render out RGB appearance using \lrasseg. We show comparisons of scene edits using SAM segments versus \lrasseg segments. \lrasseg's movable object segments consistently lead to more plausible manipulation. On the \textbf{bottom} we compare different object manipulation techniques.}
  \label{fig:object_manip_results}
  \vspace{-1.0em}
\end{figure*}


\begin{table*}[!t]
\centering
\caption{\textbf{Quantitative evaluations across various object understanding tasks.}}
\label{tab:combined}
\small
\vspace{-0.8em}
\begin{subtable}[t]{0.98\textwidth}
\centering
\caption{Point-prompted segmentation on SpelkeBench.}
\setlength{\tabcolsep}{3pt}
\begin{tabular}{lcccccccccccc}
\toprule
 & MaskFormer & SAM2 & SlotAttn & CutLER & ProMerge & ForcePrompt & PasC & EISEN & CWM & FPT & \lrasseg \\
\midrule
AR   & 0.439 & 0.482 & 0.115 & 0.321 & 0.342 & 0.051 & 0.071 & 0.158 & 0.327 &  0.368 & \textbf{0.541} \\
mIoU & 0.506 & 0.623 & 0.253  & 0.423 & 0.431 & 0.107 & 0.119 & 0.334 & 0.481 &  0.566 & \textbf{0.681} \\
\bottomrule
\end{tabular}
\label{tab:combined_pointseg}
\end{subtable}

\vspace{0.2em}

\noindent
\begin{minipage}[t]{0.48\textwidth}
  \vspace{0pt}
  \begin{subtable}[t]{\textwidth}
  \centering
  \caption{Unprompted segmentation on SpelkeBench.}
  \setlength{\tabcolsep}{4pt}
  \begin{tabular}{lccccc}
  \toprule
   & SAM2 & CutLER & ProMerge & FPT & \lrasseg \\
  \midrule
  AP   & 0.11 & 0.41 & \textbf{0.42} & 0.26 & 0.35 \\
  AR   & 0.62 & 0.32 & 0.34 & 0.18 & \textbf{0.46} \\
  mIoU & 0.68 & 0.42 & 0.43 & 0.27 & \textbf{0.57} \\
  F1   & 0.17 & 0.34 & 0.36 & 0.20 & \textbf{0.38} \\
  \bottomrule
  \end{tabular}
  \label{tab:combined_autoseg}
  \end{subtable}

  \vspace{0.2em}

  \begin{subtable}[t]{\textwidth}
  \centering
  \caption{Articulated object understanding on DragAMove.}
  \setlength{\tabcolsep}{8pt}
  \begin{tabular}{lccc}
  \toprule
   & MotionI2V & FPT & \lrasseg \\
  \midrule
  mIoU & 0.073 & 0.287 & \textbf{0.410} \\
  \bottomrule
  \end{tabular}
  \label{tab:articulated}
  \end{subtable}
\end{minipage}
\hfill
\begin{minipage}[t]{0.49\textwidth}
  \vspace{0pt}
  \begin{subtable}[t]{\textwidth}
  \centering
  \caption{Object manipulation quality on 3DEditBench.}
  \resizebox{0.77\textwidth}{!}{%
  \begin{tabular}{llccc}
  \toprule
  Method & Segment & LPIPS $\downarrow$ & SSIM $\uparrow$ & EA $\uparrow$ \\
  \midrule
  \multirow{2}{*}{PasC} & \lrasseg & \textbf{0.195} & \textbf{0.672} & \textbf{0.679} \\
                        & SAM2     & 0.241 & 0.658 & 0.536 \\
  \midrule
  \multirow{2}{*}{DH}   & \lrasseg & \textbf{0.364} & \textbf{0.555} & \textbf{0.576} \\
                        & SAM2     & 0.419 & 0.526 & 0.495 \\
  \midrule
  \multirow{2}{*}{DasS} & \lrasseg & \textbf{0.194} & \textbf{0.707} & \textbf{0.640} \\
                        & SAM2     & 0.253 & 0.682 & 0.503 \\
  \midrule
  \multirow{2}{*}{\lrasseg}
                        & \lrasseg & \textbf{0.161} & \textbf{0.736} & \textbf{0.776} \\
                        & SAM2     & 0.183 & 0.720 & 0.633 \\
  \bottomrule
  \end{tabular}%
  } 
  \label{tab:combined_objectmanip}
  \end{subtable}
\end{minipage}

\vspace{-0.4em}
\end{table*}

\begin{figure*}[!t]
  \centering
  \includegraphics[width=\linewidth]{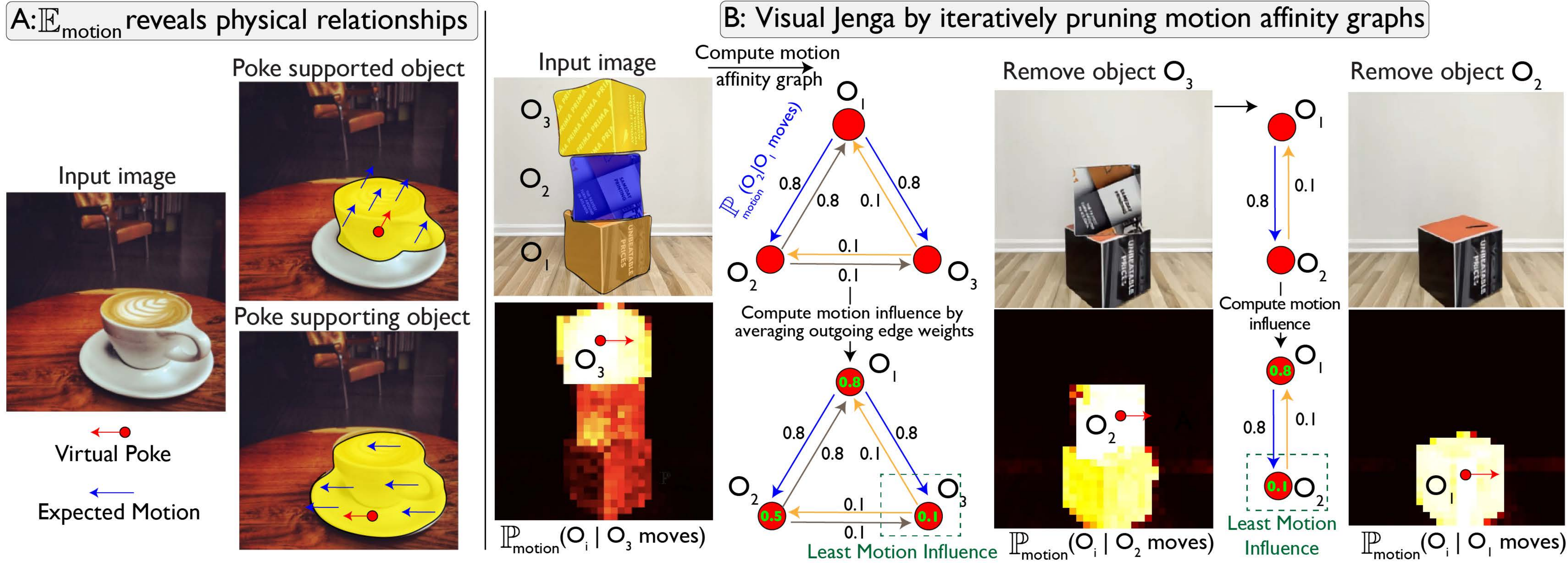}
  \vspace{-1.7em}
  \caption{
  \textbf{Reasoning about physical relationships between objects.} In \textbf{Figure A}, we show that when a virtual poke is applied to an object, the expected displacement includes not only the directly contacted object but also any objects it is physically supports. In \textbf{Figure B}, we show how probability of motion maps can be used to probe physical dependencies in scenes, enabling applications like visual jenga.} 
  \label{fig:support_relationships}
  \vspace{-1.2em}
\end{figure*}


\subsection{Unprompted movable object segmentation}
\label{sec:unprompted}
In practical settings, where systems have to operate autonomously, we may not have a point prompt on the object to begin with. They can be computed from the world model by sampling locations that are likely to move, i.e. \( p\) such that \( \mathbb{P}_{\text{motion}}[p] > \delta \), and applying the motion correlation procedure described in equation~\ref{eqn:dot_bar}. Repeating this multiple times and removing similar segments via non-maximum suppression yields a set of all movable entities in the scene. 




\textit{Results.} We evaluate on the SpelkeBench dataset and introduce two additional metrics for this task. Average Precision (AP) measures the fraction of predicted segments that end up being matched and detected, and the F1-Score balances AP and AR by computing their harmonic mean. A model that predicts only a few high quality segments may achieve high precision but low recall as it may miss many segments, while a model that over-segments may boost recall at the cost of precision. We find that \lrasseg outperforms other self-supervised methods such as CutLER~\cite{wang2023cut}, ProMerge~\cite{li2024promerge} and FPT~\cite{baumann2025if} and supervised methods like SAM2~\cite{ravi2024sam2} (see Table~\ref{tab:combined_autoseg}). Compared to supervised methods like SAM2~\cite{ravi2024sam2}, \lrasseg achieves a higher F1-score. Qualitative results shown in Figure~\ref{fig:autosegqual} suggest that SAM often over-segments scenes based on texture similarity, producing many non-physical segments which lead to poor precision and reduced interpretability for downstream physical reasoning. More results are in Supplement Section 1.6.

\subsection{Poke-prompted articulated part segmentation}
\label{sec:articulated}
Having discovered movable objects, a natural question to ask is whether there are any subparts of the object that move independently such as articulated regions. We show that given a poke on an object subpart (e.g. closing a laptop), motion correlations using the procedure described in equation~\ref{eqn:dot_bar} can identify articulated subregions of the object. We compare to FPT~\cite{baumann2025if} and MotionI2V~\cite{shi2024motion} on the DragAMove~\cite{li2024dragapart} benchmark for articulated part discovery and find that \lrasseg achieves state-of-the-art results on this task (see Table.~\ref{tab:articulated}). Additionally we show qualitative examples on the DragAMove dataset and on real images in Figure~\ref{fig:dragamove}.


\subsection{3D object manipulation}
\label{sec:manip_3d}
A crucial next step is to understand how these discovered objects move in 3D -- predicting not just what moves together, but where and how things move in space. The standard pipeline begins with an image, a point prompt and a desired 6DOF transformation as shown in Figure~\ref{fig:object_manip_results}. \lrasseg first extracts the movable object, and manipulation is achieved by specifying a dense flow field computed using the 3D transform (see Supplement Section 5.4 for more details), and generating the resulting RGB image.   

\textit{Results.} To evaluate performance of models, we use 3DEditBench~\cite{lee20253d}. For measuring performance, in addition to standard metrics like PSNR, SSIM, and LPIPS that capture image quality, we include the Edit Adherence (EA) metric introduced in prior work~\cite{pandey2024diffusion} which measures geometric edit accuracy. \lrasseg achieves state-of-the-art object manipulation performance (see Table~\ref{tab:combined_objectmanip}). Most existing methods fail on complex scenes in 3DEditBench as shown in Figure~\ref{fig:object_manip_results}. More importantly, the segments extracted from \lrasseg consistently outperform SAM, improving realism when used across diverse image editing models (PerceptionAsControl (PasC)~\cite{chen2025perception}, DiffusionHandles (DH)~\cite{pandey2024diffusion}, Diffusion-as-Shader (DasS)~\cite{gu2025diffusionshader3dawarevideo} (see Table~\ref{tab:combined_objectmanip}). SAM-generated masks capture only sub-parts of objects, resulting in fragmented or implausible edits (see Figure~\ref{fig:object_manip_results}). Additional results in Supplement Section 1.7.

\subsection{Reasoning about physical relationships}
\label{sec:support}
Understanding 3D motion tells us how individual objects behave, but many physical behaviors arise from interactions between objects. Here, we describe how \lrasseg can be used to infer pairwise physical relationships between objects. As illustrated in Figure~\ref{fig:support_relationships}a, when an object at the base of a stacked configuration is virtually perturbed, the resulting expected motion maps reveal motion across all entities it supports, providing a direct estimate of the underlying support graph. Further, probability of motion maps predicted by our world model enable computation of an object-specific \emph{movability score}, quantifying how freely an object can move without disturbing others. Formally, let the scene contain detected objects, $O_1, ... O_N $. We construct a directed graph \( G \), where each node corresponds to an object, and each directed edge from \(O_i\) to \(O_j\) has a weight \( w_{ij} = \mathbb{P}_{\text{motion}}(O_j \mid O_i \text{ moves}) \), representing the probability that object $O_j$ would move if object \(O_i\) were poked. For each object $O_i$, we compute its \emph{motion influence score}, \( \mathbb{I} [O_i] = \frac{1}{N-1} \sum_{j \neq i} w_{ij} \), by averaging the outgoing edge weights (Figure~\ref{fig:support_relationships}b). This enables tasks like Visual Jenga~\cite{bhattad2025visual} where we iteratively select the object with the least influence score, apply a large virtual poke to remove the object from the scene using the procedure in Section~\ref{sec:manip_3d}, and update the graph by pruning the corresponding node. We illustrate this procedure in Figure~\ref{fig:support_relationships}b.

\noindent






 


\section{Conclusion \& Future Work}
\label{sec:discussion}

Our work demonstrates a recipe to build a generic self-supervised, physically promptable visual world model and defines simple procedures to extract various forms of rich object understanding in a zero-shot manner, laying the groundwork for general-purpose scene understanding. Though our focus in this paper has been on human-centric macroscopic physical scenes, the underlying philosophy of using predictive models to uncover causal and structural patterns through probing could open new avenues for data-driven structure discovery in other domains where humans have less direct intuition about the nature of objecthood, such as medical imaging, astrophysics, or materials science.

\renewcommand{\thesection}{\Alph{section}}
\setcounter{section}{0}









\section*{Supplementary Materials}
\section{Overview of the supplementary} 

\begin{itemize}

    \item \textbf{Section~\ref{sec:additionalqual}: Additional qualitative results.} 
    \begin{itemize}
        \item Illustrations of parallel motion statistics --- probability of motion maps and expected motion maps (Section~\ref{sec:parallel_motion}).
        \item Illustrations of sequential generation of future world states (Section~\ref{sec:seq_gen}).
        \item Visual Jenga examples (Section~\ref{sec:visual_jenga}).
        \item Illustrations of our movable object discovery algorithm (Section~\ref{sec:addional_spelke_seg}). 
        \item Point-prompted movable object segmentation results (Section~\ref{sec:addional_pointprompt}).
        \item Unprompted movable object discovery results (Section~\ref{sec:additional_automatic}) 
        \item 3D object manipulation results (Section~\ref{sec:addional_obj_manip}).
    \end{itemize}

    \item \textbf{Section~\ref{sec:ablation_studies}: Ablation Studies.} 
    \begin{itemize}
        \item Various ablation studies: \lrasseg hyperparameter choices and scaling behavior.
    \end{itemize}
    
    \item \textbf{Section~\ref{sec:modelarch}: Additional \lrasseg architecture \& training details}  
    \begin{itemize}
        \item Details on \lrasseg architecture, quantizer design, and training procedures.
    \end{itemize}

    \item \textbf{Section~\ref{dataset_details}: SpelkeBench dataset details}  
    \begin{itemize}
        \item Dataset collection procedure and more visual examples from the dataset.
    \end{itemize}

    \item \textbf{Section~\ref{sec:inference_procedures}: Details about \lrasseg's applications.}
    \begin{itemize}
        \item Further details on procedures that enable \lrasseg's applications like computation of parallel motion statistics, segment extraction and 3D object manipulation.
    \end{itemize}
    
    \item \textbf{Section~\ref{sec:baseline_supp}: Baseline evaluation details, and additional qualitative comparisons.} 
    \begin{itemize}
        \item Here we show the settings used for point-prompted movable object segmentation baseline evaluations, and present additional qualitative analysis showing failure modes.
    \end{itemize}
    \item \textbf{Section~\ref{sec:stat_sig}: Statistical significance of results} 
    \begin{itemize}
        \item In this section, we present some analysis of statistical significance of the quantitative results on various object understanding applications in the main paper.
    \end{itemize}

\end{itemize}

\section{Additional Qualitative Results}
\label{sec:additionalqual}

\subsection{Parallel motion statistics}
\label{sec:parallel_motion}

In the main paper, we introduced the use of parallel motion statistics to summarize how different regions of a scene are likely to move under interaction, and we presented qualitative examples in Figure 2. Here, in Figure~\ref{fig:supp_pmotion}, we provide additional examples spanning a wider range of scenes and objects. These qualitative results further demonstrate the consistency of the predicted motion probabilities and the model’s ability to capture coherent motion statistics across diverse visual environments.

\subsection{Sequential generation of future world states}
\label{sec:seq_gen}

In the main paper (Figure 3), we demonstrated the model's ability to generate multiple plausible future world states through sequential autoregressive rollouts. These examples highlighted both the uncertainty in object trajectories and the physical coherence maintained across generations. Here, in Figure~\ref{fig:supp_uncond}, we present additional qualitative results spanning a more diverse set of objects, interactions, and scene configurations. These further illustrate the robustness of the model’s temporal predictions and its capacity to synthesize physically consistent futures.

\subsection{Visual Jenga examples}
\label{sec:visual_jenga}

In the main paper, we applied our world model to the Visual Jenga task and presented qualitative examples in Figure 6b. These results demonstrated the model's ability to identify objects which can be freely moved without disturbing other objects in the scene. We showed that our world model can unstack a stacked structure iteratively while maintaining structural stability in each step. Here, in Figure~\ref{fig:support_relationships}, we provide additional Visual Jenga examples that cover a broader set of tower configurations and intervention types. These results further highlight the model’s capacity to reason about physical relationships in complex scenes.

\subsection{Illustrations of our movable object discovery algorithm}
\label{sec:addional_spelke_seg}
In Section 4.3 and Figure 4a of the main paper, we outlined our algorithm for discovering movable object segments by simulating virtual pokes and computing correlated motion statistics in the flow responses. Here, in Figure~\ref{fig:point_seg_process}, we provide additional examples that visualize the full process—from poke point sampling to segment extraction. These visualizations further demonstrate the robustness and consistency of our flow-based grouping method across a diverse range of objects and scenes.

\begin{table*}[t!]
  \centering
  \caption{\textbf{Ablation studies.} We analyze the effect of modifying the parameters used in the segment discovery algorithm described in Section 4.3 of the main paper, impact of model scaling, SAM2's mask selection strategy, and the importance of incorporating flow tokens for training \lrasseg. These experiments are conducted on the point-promoted movable object segmentation task on \spelkeentity.}
  \label{tab:merged}
  \small
  \setlength{\tabcolsep}{4pt}
  \begin{tabular}{cc}
    \begin{subtable}[t]{0.48\linewidth}
      \centering
      \caption{\textbf{Parameters used in the object discovery algorithm}}
      \label{tab:statprobingablation}
      \begin{tabular}{lcccc}
        \toprule
         & \multicolumn{4}{c}{\#seeds=1, \#sequential steps=64} \\
         \cmidrule{2-5}
        \#pokes & 1 & 2 & 4 & 8  \\
        \midrule
        AR   & 0.3793 & 0.3938 & 0.4673 & \textbf{0.5251}  \\
        mIoU & 0.5874 & 0.5964 & 0.6419 & \textbf{0.6786}  \\
        \toprule
        & \multicolumn{4}{c}{\#pokes=1, \#sequential steps=64} \\
         \cmidrule{2-5}
        \#seeds & 1 & 2 & 4 & 8  \\
        \midrule
        AR   & 0.3793 & 0.3987 & 0.4495 & \textbf{0.4822} \\
        mIoU & 0.5874 & 0.5943 & 0.6266 & \textbf{0.6448} \\
        \toprule
        & \multicolumn{4}{c}{\#seeds=1, \#pokes=8} \\
         \cmidrule{2-5}
        \#sequential steps & 0 & 64 & 128 & 256 \\
        \midrule
        AR   & 0.4622 & 0.5251 & 0.5314 & \textbf{0.5336} \\
        mIoU & 0.6413 & 0.6715 & 0.6750 & \textbf{0.6774} \\
        \bottomrule
      \end{tabular}
    \end{subtable}
    &
    \begin{subtable}[t]{0.48\linewidth}
      \centering
      \caption{\textbf{Scaling behavior}}
      \label{tab:scalingablation}
      \begin{tabular}{lccc}
        \toprule
        & \multicolumn{3}{c}{\#pokes=8, \#seeds=1, \#sequential steps=64} \\
         \cmidrule{2-4}
        & \lrasseg (100M) & \lrasseg (1B) & \lrasseg (7B)  \\
        \midrule
        AR   & 0.4306 & 0.5251 & \textbf{0.5466} \\
        mIoU & 0.6166 & 0.6715 & \textbf{0.6804} \\
        \bottomrule
      \end{tabular}
      \vspace{0.5em}
      \caption{\textbf{SAM2 mask selection ablations}}
      \label{tab:sam2ablation}
      \begin{tabular}{lccc}
        \toprule
        & Most confident & Random & Least confident  \\
        \midrule
        AR   & \textbf{0.4816} & 0.4070 & 0.3025  \\
        mIoU & \textbf{0.6225 }& 0.5900 & 0.5012  \\
        \bottomrule
      \end{tabular}
      \vspace{0.5em}
      \caption{\textbf{Importance of using flow tokens}}
    \label{tab:rgbvsflow}
    \begin{tabular}{lccc}
      \toprule
      & CWM & \lrasseg-RGB & \lrasseg  \\
      \midrule
      AR   & 0.158 & 0.412 & \textbf{0.541} \\
      mIoU & 0.334 & 0.576 & \textbf{0.681} \\
      \bottomrule
    \end{tabular}
    \end{subtable}
    \\
  \end{tabular}
\end{table*}

\subsection{Point-prompted movable object segmentation results}
\label{sec:addional_pointprompt}
In Section 4.3 of the main paper, we evaluated segmentation quality under point-prompted settings on \spelkeentity. Here, we present additional qualitative results comparing our method to several baselines. As shown in Figure~\ref{fig:point_seg_supplementary}, our method consistently produces cohesive and physically plausible segments, in contrast to alternative approaches that often fragment objects or include extraneous background elements.

\subsection{Unprompted movable object segmentation results}
\label{sec:additional_automatic}
Previously, in Figure 4c of the main paper, we evaluated unprompted segmentation on the \spelkeentity benchmark. Here, we present in Figure~\ref{fig:auto_seg_supplementary} additional qualitative results comparing our method to baselines such as SAM2~\cite{ravi2024sam2}, ProMerge~\cite{li2024promerge}, and CutLER~\cite{wang2023cut}, and FPT~\cite{baumann2025if}. As shown, our method consistently discovers a set of physically plausible segments, whereas the baselines tend to over-segment or under-segment the scene.

\subsection{3D object manipulation results}
\label{sec:addional_obj_manip}
In Section 4.5 of the main paper, we demonstrated \lrasseg's ability to manipulate objects in 3D, and the importance of physically grounded segmentation for object manipulation. Here, in Figure~\ref{fig:suppl_obj_manipulation} we include further qualitative comparisons of edits generated using our predicted segments versus those from SAM2~\cite{ravi2024sam2} and additional comparisons with SOTA 3D editing methods in Figure~\ref{fig:suppl_obj_manipulation_comp}. As illustrated, segments aligned with physical objecthood based on co-movement significantly improve edit realism, spatial coherence, and transformation consistency across multiple \spelkeeditbench~\cite{lee20253d} examples.

\section{Ablation studies} 
\label{sec:ablation_studies}
We conduct a series of ablations to assess the effect of various design choices. All experiments are done on the point-prompted movable object segmentation application described in Section 4.3 of the main paper. 
\begin{itemize}
    \item Table~\ref{tab:statprobingablation} validates the segment extraction framework detailed in Section 4.3 of the main paper: averaging across multiple pokes and seeds improves and stabilizes performance. The table also compares parallel decoding (0 steps) with sequential decoding using varying numbers of autoregressive steps (64-256), a process described in detail at the end of Section 3 of the main paper. Results show that while sequential decoding improves upon parallel mode, performance gains diminish beyond 64 steps, suggesting that most significant causal dependencies can be captured with relatively few steps.
    \item Table~\ref{tab:scalingablation} shows that scaling the model is beneficial up to the 7B parameter range. 
    \item Table~\ref{tab:sam2ablation} compares the mask selection strategies for running SAM2 evaluations on \spelkeentity in multimask mode. Specifically, SAM2's multimask mode outputs a set of masks and their associated confidence score. To give SAM2 the fairest chance on the benchmark, we perform an ablation study on mask selection strategy, including most confident, random, and least confident. We find that choosing the most confident mask outperforms random or least confident selection strategies, but still falls short of our approach. 
    \item Table~\ref{tab:rgbvsflow} shows the importance of using flow tokens by comparing \lrasseg to models trained without flow tokens. We evaluate CWM\cite{bear2023unifyingmachinevisioncounterfactual}, \lrasseg, and \lrasseg-RGB—a variant trained to operate in RGB space rather than flow space (following the same training methodology described in Section 3 of the main paper except without the flow tokens). For \lrasseg-RGB, we use the same RGB patch motion counterfactual method proposed in CWM for extracting plausible flow fields. While \lrasseg-RGB outperforms CWM due to its multimodal generative capabilities, \lrasseg with flow tokens substantially outperforms the RGB-only variant, as flow provides a more intuitive control surface for exercising virtual physical interventions in the scene.
\end{itemize}

\section{Additional \lrasseg specifications}
\label{sec:modelarch}

\textbf{Pointer tokens enable random order sequence constructions, making probabilistic models tractable.}
\label{sec:pointer_content_data}
\lrasseg adapts the causal autoregressive modeling paradigm for high-dimensional data. Traditional GPT-style transformers predict sequences in a preset, hard-coded order. While this is a natural fit for one-dimensional data such as language, it becomes an unnecessary, and potentially harmful inductive bias when modeling higher-dimensional data. Most autoregressive image modeling approaches simply accept this bias, while we introduce a new token type---the pointer---which allows us to serialize the data in arbitrary order.

Pointer tokens enable us to package random-access traversals over high-dimensional data structures (such as images) into one-dimensional sequences of tokens for efficient causal pretraining, by interleaving pointer tokens among the content tokens, which represent the actual data, as illustrated in Figure 1b in the main paper.


In addition to freeing us from the raster-order generation bias, pointer tokens allow us to condition our model on any subset of the image, learning complex multidirectional conditioning relationships in the data, essentially learning the underlying probabilistic model (PGM) of the world. They also allow for partial patch conditioning, and patch regeneration during inference. All of this functionality can still be simply expressed as a causal autoregressive sequence of tokens, and thus can be modeled and optimized as a standard LLM.

The sequence model formulation transforms the challenge of learning probabilistic models at scale. The key insight is that our sequences represent traversals through the probabilistic model---each pointer-content pair corresponds to visiting and observing a node in the graph. By modeling these traversals autoregressively, we approximate the full joint distribution through conditional factorization. The pointer mechanism ensures we can still query any conditional distribution at spatiotemporal locations, but now through tractable sequential prediction rather than exponentially complex full inference. This recasts the seemingly intractable problem of learning a complete probabilistic model over high-dimensional visual data as a standard GPT-style modeling problem.

\textbf{Our learned local quantizer.}
\label{sec:hlq}
Most visual autoregressive models utilize popular off-the-shelf quantizers such as VQ-GAN~\cite{esser2021taming}, VQ-VAE~\cite{van2017neural}, or the Cosmos tokenizer~\cite{agarwal2025cosmos}. While such standard quantizers achieve strong compression ratios, they do not preserve locality of the patches within the token space, but rather encode the image in its entirety as a global code. While some locality is certainly present in the token representation, swapping any given token can modify the representations associated with patches on the other side of the image. While it is naturally more efficient to compress information globally, this comes at a cost of a less interpretable and less controllable latent (token) space.

Instead of compressing the whole frame into global codes, the \lrasseg architecture uses a Hierarchical Local Quantizer (HLQ): a convolutional autoencoder whose receptive field never crosses patch boundaries during encoding. Each 16 × 16 RGB pixel patch is encoded into a sequence of four 16-bit codes using finite scalar quantization~\cite{mentzer2023finite}, yielding a 65,536-token vocabulary for RGB images. A second, similar quantizer is used to quantize optical-flow patches. No information from neighboring patches leaks into a code, preserving strict locality. This lets local downstream interventions---masking, overwriting, or re-ordering individual patches---behave predictably. Additionally, it makes the autoregressive modeling objective better aligned with natural-language modeling assumptions of token independence. A decoder accepts local 16-bit codes at every spatial location, and decodes them jointly into the input image/flow map that produced them.

\textbf{Dataset preparation.}
The world model is pre-trained on BVD (Big Video Dataset~\cite{lee20253d})---a 7,000 hour dataset of diverse Internet videos mixed with standard 3-D vision datasets such as ScanNet++ \cite{yeshwanth2023scannetpp}, CO3D \cite{reizenstein2021co3d}, RealEstate-10K \cite{zhou2018realestate10k} and standard video datasets such as Kinetics \cite{kay2017kinetics}, SomethingSomethingv2 \cite{goyal2017something} and OpenX embodiment~\cite{open_x_embodiment_rt_x_2023}. Camera pose information is provided to the model whenever available in the dataset, and optical flow for every frame pair is computed with the SeaRAFT \cite{wang2024searaft} model and quantized. The BVD also consists of internet videos automatically crawled using search queries generated by LLaMA~3 \cite{Dubey2024TheL3}.  The queries targeted videos containing rich physical dynamics, diverse environments, and varied objects. Specifically, action categories from Kinetics400 \cite{kay2017kinetics} were expanded with additional sports, physical activities, and product review categories. To ensure training relevance, we filtered videos by requiring a minimum level of optical flow and by applying CLIP \cite{radford2021clip}-based keyword alignment. Positive keywords included \emph{action}, \emph{activity}, \emph{motion}, and \emph{place}, while negative keywords included \emph{animation}, \emph{cartoon}, \emph{face}, \emph{game menu}, \emph{graphic}, \emph{map}, \emph{newscast}, \emph{person}, and \emph{screenshot}. Alignment was quantified by the dot product between CLIP embeddings of keywords and video frames.

\textbf{Key Training Details.}
We train an 80M-parameter RGB HLQ on a combination of ImageNet and Open Images, and an 80M Flow HLQ on the BVD dataset. We quantize 512x512 RGB images and flow maps into tokens and train a 7B-parameter $\Phi$ model on a dataset of 3 million RGB video clips. This sums up to about 1.4 trillion tokens. We train $\Phi$ with causal sequences of 2 frames, spanning up to 1 second of video. Mixed-precision training on 64 H100 GPUs at 65\% MFU yields 490 TFLOPS/device ($\sim$31 PFLOPS total) sustained.

All models were trained autoregressively using cross-entropy loss on next-token prediction with a batch size of $512$. We first train with only RGB and camera pose tokens for $5\times10^{5}$ steps under a Warmup-Stable-Decay (\textbf{WSD}) schedule. The learning rate was linearly warmed up over $2{,}000$ iterations to $3\times10^{-4}$, held constant until the final $1\times10^{4}$ steps, and then decayed linearly to zero. Training was continued for an additional $2\times10^{5}$ steps with optical flow tokens, in addition to RGB and camera tokens, where the learning rate was re-warmed over $800$ iterations to $3\times10^{-4}$, maintained at this value, and decayed linearly to zero during the last $1\times10^{4}$ steps. Each training step takes approximately 3.8s. It has been shown that WSD achieves similar or better performance than cosine schedules of the same length. The use of WSD is particularly important here not because of superior performance, however, but because it enables continual training in the flow tokens integration step we discussed above.

\begin{figure}[!t]
  \centering
  \includegraphics[width=0.7\linewidth]{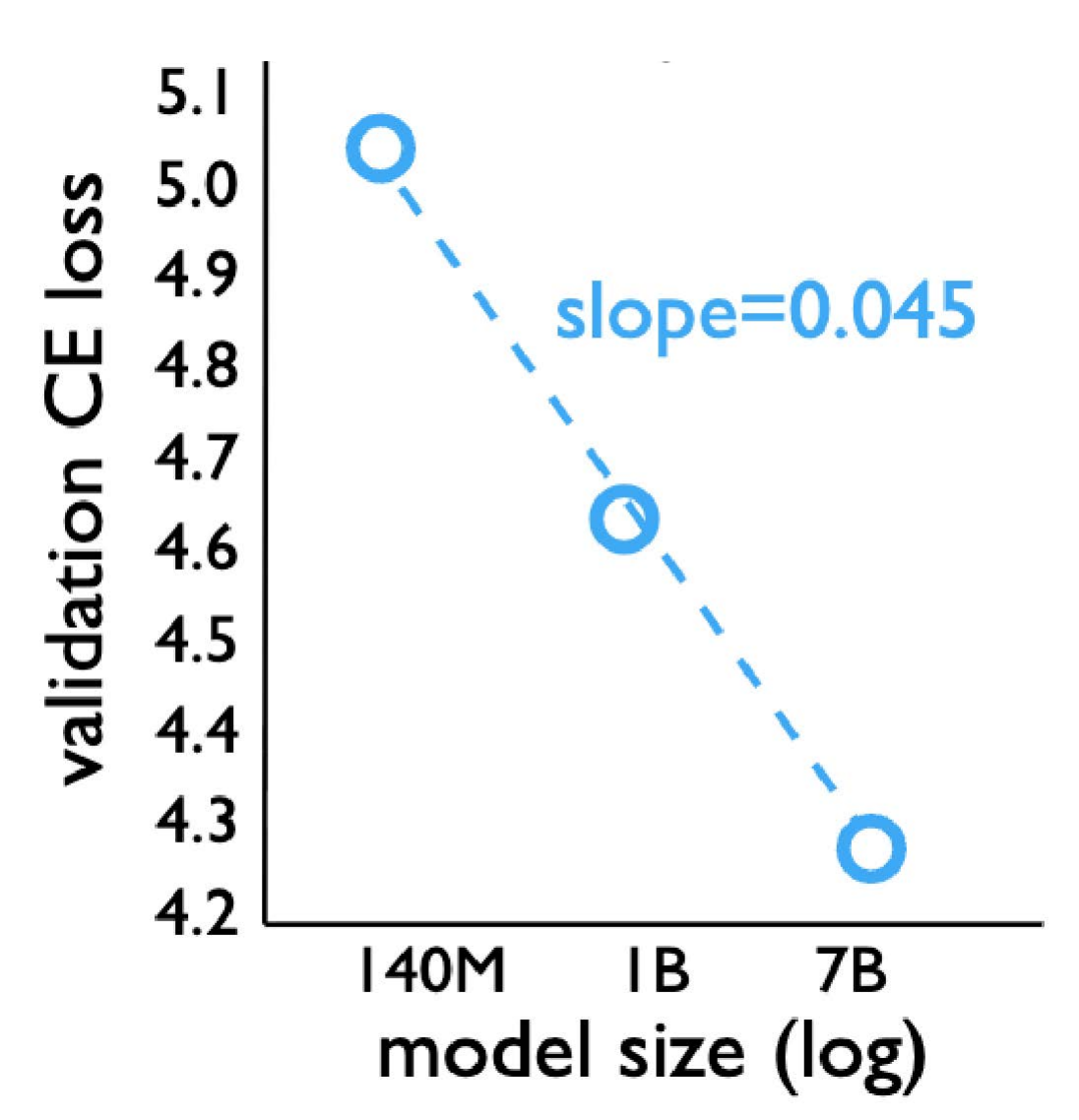}
  \caption{
  \textbf{Scaling laws that show our model obtaining lower loss when given more parameters.} }
  \label{fig:scaling}
\end{figure}

\textbf{Scaling Properties of \lrasseg.} A critical advantage of the \lrasseg architecture is its predictable scaling behavior, inheriting the well-established scaling laws of language models while extending them to visual domains. We trained \lrasseg models from 100M to 7B parameters and observed consistent improvements in validation loss across three orders of magnitude (Figure~\ref{fig:scaling}). The reliable scaling without saturation even at 7B parameters suggests that further scaling would yield continued benefits, validating that \lrasseg successfully transfers the scaling properties of autoregressive language models to structured visual data. We also show scaling properties on downstream applications like point prompted segmentation in our ablation studies here in Section~\ref{sec:ablation_studies}.

\section{\spelkeentity dataset details}
\label{dataset_details}

Here we provide more details on SpelkeBench~\cite{SpelkeBench} --- the evaluation benchmark we use for movable object discovery in Sections 4.3 and 4.4 of the main paper. This benchmark is designed to assess whether segmentation algorithms can identify movable objects --- defined as regions that move together --- unlike existing datasets such as COCO~\cite{lin2015microsoftcococommonobjects}, which emphasize semantic or instance labels. We expand on collection procedures and provide qualitative examples from the dataset. 

\subsection{Collection procedure}
\spelkeentity curates a dataset from two complementary sources: the EntitySeg benchmark~\cite{Qi_2023_EntitySeg} and the OpenX-Embodiment robotics dataset~\cite{open_x_embodiment_rt_x_2023}. While EntitySeg is designed for high-resolution internet imagery with dense segmentation annotations, OpenX consists of real-world, egocentric robot interactions. Since OpenX does not provide segment labels, they are manually annotated for a subset of 50 images. These annotations reflect the types of movable objects relevant for physical interaction and manipulation tasks that are central to robot learning. For EntitySeg, a high-quality subset of 500 images is extracted using a three-stage filtering pipeline that filters out the annotated segments in the dataset which do not align with the definition of movable objects:

\begin{itemize}
    \item \textit{Stage 1: Removal of amorphous background regions.} All regions labeled as ``stuff''—such as sky, ground, or terrain—based on the standard stuff-vs-things taxonomy~\cite{Qi_2023_EntitySeg} are excluded. These regions lack the individuated, cohesive properties associated with movable objects and are typically not physically manipulable entities.

    \item \textit{Stage 2: Filtering non-movable object categories.} Despite being labeled as ``things'', certain objects like kitchen sinks, traffic signs, or large fixtures are functionally immovable in real-world settings. These are identified and removed through manual inspection.

    \item \textit{Stage 3: Final curation of diverse, high-quality scenes.} From the filtered pool, 500 images that contain only movable object-consistent annotations are selected. It is ensured that this set is diverse in terms of object types, spatial arrangements, and scene complexity.
\end{itemize}

\subsection{\spelkeentity qualitative examples}
\label{sec:spelkebench_suppl}
 In Figure~\ref{fig:entityseg_dataset_suppl}, we present additional examples of images and associated ground-truth movable object annotations in SpelkeBench. These examples further show the failures of existing segmentation benchmarks such as EntitySeg~\cite{Qi_2023_EntitySeg} and SA-1B~\cite{ravi2024sam2}, which often contain segments that diverge from the movable object criteria, such as amorphous background regions, or subregions of objects. SpelkeBench's physically grounded segment annotations are therefore better candidates to evaluate the object discovery applications of \lrasseg presented in the main paper. 

\section{Details about \lrasseg's applications}
Here we provide additional details for the algorithms that enable various object understanding applications of \lrasseg.
\label{sec:inference_procedures}

\subsection{Parallel estimation of motion statistics}

In Section 4.1 of the main text, we introduced the expected motion map as the probability weighted average of flow vectors that map to flow tokens. This flow token to flow vector mapping needs to be done as a necessary step before the expectation can be computed. Here we describe the procedure used for doing that. 

\noindent \textbf{Flow token epigraphy.} 
\label{sec:epigraphy}
\lrasseg uses a learnt \emph{local patch quantization} to produce flow tokens, but relies on a \emph{global decoder} to generate coherent, high-quality flow fields. As a result, tokens cannot be interpreted by decoding them in isolation---their meaning emerges only in the context of the full sequence. However, since the tokenizer is local, we can find which continuous flow vectors map to it by performing a kind of token space epigraphy---by assigning meaning to discrete flow tokens through statistical aggregation of typical input flow fields that produced them:
\[
\begin{aligned}
f_j \!\mapsto\!  \mathbf{v}_j &= \frac{1}{|S_j|} \sum_{\mathbf{u} \in S_j} \mathbf{u}, \\
\quad \text{where,}\quad  S_j &= \left\{ \mathbf{u} \in \mathbb{R}^2 \mid \text{tokenizer}(\mathbf{u}) = f_j \right\}.
\end{aligned}
\]

\subsection{Movable segment discovery procedure} 
\label{sec:stat_probing_intuition}
As discussed in Section 4.3 of the main paper, we use \lrasseg for movable object discovery by computing the statistical aggregate of pixel-to-pixel correlated motion statistics across a variety of applied virtual pokes and generation seeds. In this section, we explain the intuition behind the process and how it allows for a more expressive definition of movable objects.

We build upon the idea of counterfactual probing introduced in CWM~\cite{bear2023unifyingmachinevisioncounterfactual}, where movable objects are discovered by simulating localized virtual pokes through local patch motion interventions and analyzing the outcome. However, due to its regression-based nature, CWM produces a single deterministic output, which in practice does not accurately represent the physical world where there are multiple physically plausible outcomes for a poke. Consider a simple example of moving a person's hand, the hand may move independently of the body or the entire body may move with the hand--both being valid outcomes. However, because of the aforementioned deterministic nature of CWM, it is forced to average over these distinct possibilities, leading to ambiguous or blurry motion completions that fail to reveal which parts of a scene tend to move together.

In contrast, because of the generative nature of \lrasseg which generates multiple plausible future motions of a scene, we can operationalize movable objects with a more expressive definition as groups of pixels that consistently move together across multiple plausible outcomes of a world model, under different virtual pokes. This requires modeling the distribution of possible responses to external forces.
As such, the algorithm defined in Section 4.3 of the main paper is a natural stochastic extension of the original CWM counterfactual procedure as instead of a single prediction, we use \lrasseg to produce a diverse set of imagined flow completions for various virtual pokes at a candidate spatial location. Diversity arises from two sources of randomness:
\begin{enumerate}
    \item \textbf{Sampling flow tokens} from the learned distribution $\text{Pr}[v | \mathbf{X} \circ p ]$ at a pointer $p$: we draw multiple flows $f_k \sim \text{Pr}[v | \mathbf{X} \circ p ]$ to explore the local responses the model deems plausible (e.g. if an object is on a table, we would not sample pokes down into the table).
    \item \textbf{Varying the decoding order} of spatial indices $p_k$: Because \lrasseg is a sequence model, tokens decoded earlier condition those decoded later. Shuffling the order therefore changes how motion propagates through the object---e.g. decoding the torso \emph{before} the leg yields a different global outcome than decoding the leg first.
\end{enumerate}

In our evaluations, we use 8 pokes, and also perform a refinement step by zooming into the object with the initial estimate of the segment, and repeating the extraction procedure.

\subsection{Unprompted segmentation procedure}

\label{sec:autoseg_procedure}
Here we provide additional details about our unprompted segmentation algorithm described in Section 4.4 in the main paper. In many real-world settings, especially in robotics, it is advantageous to automatically discover \textit{every} independently movable object in a scene without requiring manual point-prompting. For example, a household robot tasked with clearing a dining table must infer that a plate and its contents will move as a unit, while a napkin resting on the plate is an independent entity, so it can plan appropriate grasps and avoid unintended collisions. 

We now describe a method to extract the full set of movable object segments in a scene automatically. Our approach consists of two steps. First, as described in Section 4.4 we poke the scene at multiple locations sampled from the probability of motion map. Then from the model's flow responses, we compute a dense pixel-to-pixel affinity matrix that captures the likelihood that a pair of pixels will move together. An iterative clustering algorithm is applied to this matrix to isolate a complete set of independently movable entities. 

\textit{Computing the affinity matrix}. We begin by sampling locations from the motion probability map. These points are where we “poke” to collect flows.
\[
\mathcal{K} \;=\;\{\,p_{1},\,p_{2},\,\dots,\,p_{N}\}
\;\subset\;\mathcal{I},
\qquad
\mathbb{P}_{\mathrm{motion}}[p_i]\;>\;\tau_{p}.
\]

We then build a motion descriptor for each pixel using the following procedure:

For each \(n=1,\dots,N\), choose \(R\) poke‐directions \(\{f_n^{(r)}\}_{r=1}^R\).  For each \((n,r)\) and each of \(t=1,\dots,T\) random seeds, compute the flow completion given the input image tokens $\mathbf{r}^0$,
\[
\hat{\mathbf{f}_t}^{(n, r)} \overset{\text{seq}}{\sim}\Phi(\mathbf{r}^0 \circ  [c=0] \circ [p_{n}, f_n^{(r)}]; \text{flow}; \texttt{seed}=t)
\]
Then for each \(u\in\mathcal{I}\), where \(\mathcal{I}\) is the set of 2D pixel locations, the motion descriptor,
\[
\varphi[u] =\bigl[\hat{\mathbf{f}_1}^{(1, 1)},\,\dots,\,\hat{\mathbf{f}_t}^{(n, r)}(u)\bigr]
\;\in\;\mathbb{R}^{2\,N\,R\,T}.
\]
Finally, the affinity matrix can be described as the pairwise dot product of motion descriptors: 
\[
A[u,v]=\varphi[u]^\top\,\varphi[v],
\quad
\forall\,u,v\in\mathcal{I}.
\]
For simplicity, we denote $A[u]$ to be the affinity of the pixel $u$ with the rest of the image. 

\textit{Clustering the affinity matrix to extract segments.}  Given the precomputed affinity matrix \(A\), we extract segments in an iterative “select–threshold–refine” loop. At each step, we choose the most confident probe center \(k_{i^*}\), defined as the one whose affinity‐row \(A[k_{i^*}]\) has the highest mean over all pixels---indicative of strong binding to the other pixels that make up the object. We apply Otsu’s method to threshold this row, yielding an initial mask \(M^{(0)}\). We then gather all remaining poke points \(k_j\) that lie within \(M^{(0)}\) and average their affinity‐rows to form:
\[
A_{\mathrm{avg}}
= \frac{1}{\lvert\{j: k_j\in M^{(0)}\}\rvert}
  \sum_{k_j \in M^{(0)}} A[k_j]
\]
We threshold \(A_{\mathrm{avg}}\) via Otsu's method to obtain a refined mask \(M ^ {(t)}\), for $t=0$. All centers contained in \(M^{(t)}\) are then removed from consideration, and the loop repeats on the remaining set of poke points. Once no poke points remain, the algorithm returns the complete set of extracted segments \(\{M^{(1)},\dots,M^{(T)}\}\). Non-maximum suppression is then used to remove duplicate segments.  Figure~\ref{fig:autosegschematic} illustrates this procedure using an example.

\begin{table*}[t]
\centering
\small
\resizebox{\linewidth}{!}{%
\begin{tabular}{lllccc}
\toprule
\textbf{Task} & \textbf{Baseline} & \textbf{Metric} & $\boldsymbol{\Delta \pm \text{SE}}$ & \textbf{95\% CI} & \textbf{$p$-value} \\
\midrule
 \multirow{2}{*}{\textbf{Point Segmentation}} & \multirow{2}{*}{\begin{tabular}[c]{@{}l@{}}
\lrasseg vs SAM2 \\
\textit{Tab. 1a (col 12 vs col 3)}
\end{tabular}} 
& mIoU $\uparrow$  & $+0.058 \pm 0.010$ & $[0.038,\;0.077]$ & $<0.001$ \\
& & AR  $\uparrow$  & $+0.059 \pm 0.013$ & $[0.034,\;0.084]$ & $<0.001$ \\
\midrule
\multirow{4}{*}{\textbf{Automatic Segmentation}} & 
\multirow{4}{*}{\begin{tabular}[c]{@{}l@{}}
\lrasseg vs ProMerge \\
\textit{Tab. 1b (col 6 vs col 4)}
\end{tabular}}
& mIoU $\uparrow$ & $+0.14 \pm 0.01$ & $[0.12,\;0.16]$  & $<0.001$ \\
& & AP $\uparrow$   & $-0.07 \pm 0.01$ & $[-0.10,\;-0.05]$ & $<0.001$ \\
& & AR $\uparrow$   & $+0.12 \pm 0.01$ & $[0.09,\;0.14]$  & $<0.001$ \\
& & F1 $\uparrow$   & $+0.02 \pm 0.01$ & $[0.00,\;0.04]$  & $0.082$ \\
\midrule
\multirow{6}{*}{\textbf{Object Manipulation}} 
& \multirow{3}{*}{\begin{tabular}[c]{@{}l@{}}
\lrasseg vs PasC \\
\textit{Tab. 1d (row 8 vs  row 2)} \\
\end{tabular}}
& EA $\uparrow$  & $+0.096 \pm 0.030$ & $[0.044,\;0.150]$ & $<0.001$ \\
& & LPIPS $\downarrow$ & $-0.035 \pm 0.009$ & $[-0.041,\;-0.028]$ & $<0.001$ \\
& & SSIM $\uparrow$ & $+0.064 \pm 0.006$ & $[0.049,\;0.081]$ & $<0.001$ \\
\cmidrule{2-6}
& \multirow{3}{*}{\begin{tabular}[c]{@{}l@{}}
\lrasseg seg. vs SAM2 seg. \\
\textit{Tab. 1d (row 8 vs  row 9)}
\end{tabular}}
& EA $\uparrow$  & $+0.143 \pm 0.042$ & $[0.064,\;0.223]$ & $<0.001$ \\
& & LPIPS $\downarrow$ & $-0.022 \pm 0.007$ & $[-0.028,\;-0.017]$ & $<0.001$ \\
& & SSIM $\uparrow$ & $+0.016 \pm 0.013$ & $[-0.015,\;0.047]$ & $0.355$ \\
\bottomrule
\end{tabular}
}
\vspace{-1em}
\caption{Statistical analysis of relative improvement ($\Delta$) of \lrasseg vs the second best model across three tasks.}
\label{tab:merged_results}
\vspace{-1em}
\end{table*}

\subsection{3D Object Manipulation Procedure}
\label{sec:supp_3d_manip}
As discussed in Section 4.5 in the main paper, \lrasseg achieves state-of-the-art object manipulation performance by leveraging 2D optical flow fields that encode 3D transformations. Specifically, to perform 3D object manipulation, we create a flow field where the flow on the surface of the object characterizes the 3D transformation to be performed, with the flow of the background set to 0. We use this to condition the predictor to move the object, but keep the background fixed. To generate these flow fields, we perform the following procedure: 
\begin{itemize}
    \item Step 1: Unproject the depth map of the input image, obtained using off-the-shelf supervised metric depth estimator Depth Anything V2~\cite{yang2024depthv2}, to obtain a 3D point cloud.
    \item Step 2: Apply the desired rigid transformation to the object's point cloud while keeping background points unchanged.
    \item Step 3: Re-project the transformed point cloud and compute the displacement relative to the original pixel positions to generate the 2D flow field.
    \item Step 4: Finally, \lrasseg then generates the manipulated image given the computed object-masked flow map and the input image.
\end{itemize}
\section{Baseline evaluation details and additional qualitative comparisons}
\label{sec:baseline_supp}
In Section 4.3 of the main paper, we evaluate various baselines on \spelkeentity using point-prompted segmentation, wherein each method receives a poke point and must output a binary segment. In this section, we expand on how  baseline methods generate binary segments from single-point prompts for evaluation, and present additional qualitative analysis showing failure modes of these methods. 
\begin{itemize}
    \item \textbf{SAM2~\cite{ravi2024sam2}}: We use single-point prompting on SAM2 multimask mode and choose the most confident mask (see ablation in Table~\ref{tab:sam2ablation}).
    \item \textbf{DINOv2~\cite{oquab2023dinov2}}: We extract features using the ViT-G/14 backbone, compute affinity between the poke point's feature and all spatial features, threshold using Otsu's method, and apply Conditional Random Field refinement to obtain the final binary mask.
    \item \textbf{CutLER~\cite{wang2023cut} \& ProMerge~\cite{li2024promerge}}: Both methods generate multiple candidate segments. We select the segment (if any) containing the poke point.
    \item \textbf{ForcePrompting~\cite{gillman2025force},  PerceptionAsControl~\cite{chen2025perception}}: For these drag-based approaches, we use the poke point to initialize a trajectory to generate the edited image. We then compute optical flow between original and edited image using RAFT and apply Otsu thresholding to the flow magnitude to extract the binary segment. Some failure modes of these models are illustrated in Figure~\ref{fig:perceptionascontrol} 
    \item \textbf{FPT~\cite{baumann2025if}}: FPT produces a flow completion given a virtual poke, similar to our method. We can directly apply our algorithm in Section 4.3 of the main paper to this method. 
    \item \textbf{CWM~\cite{venkatesh2024understanding}}: Similar to drag-based approaches, we apply interventions at the poke point, compute RAFT flow between original image and intervention outcome, and threshold the flow magnitude to obtain the binary segment. CWM merges nearby objects because the model often generates blurry reconstructions, as its RGB pixel regression objective during training does not account for uncertainty. As a result, the flow estimation may produce diffuse or extended motion fields, causing nearby objects to be grouped together, as illustrated in Figure~\ref{fig:cwmcomparisonsuppl} 
\end{itemize}

\section{Statistical significance of quantitative results} 
\label{sec:stat_sig}

To ensure that our improvements are not only numerically higher but also statistically reliable, we perform a comprehensive significance analysis across all quantitative evaluations. In particular, we focus on the tasks reported in Table 1a, 1b, and 1d of the main paper. Specifically,  we report the relative improvement ($\Delta$), along with standard errors (SE), confidence intervals (CIs), and paired t-tests. Across nearly all metrics in Table~\ref{tab:merged_results}, we observe three consistent trends: (1) the standard errors are small, indicating low variance across samples; (2) the confidence intervals are tight and do not cross zero, suggesting that the observed improvements are consistently positive; and (3) paired t-tests confirm statistical significance with $p \leq 0.001$. Taken together, these results demonstrate that our improvements reflect stable and statistically significant gains over prior methods.

\begin{figure*}
  \centering
  \includegraphics[width=0.98\textwidth]{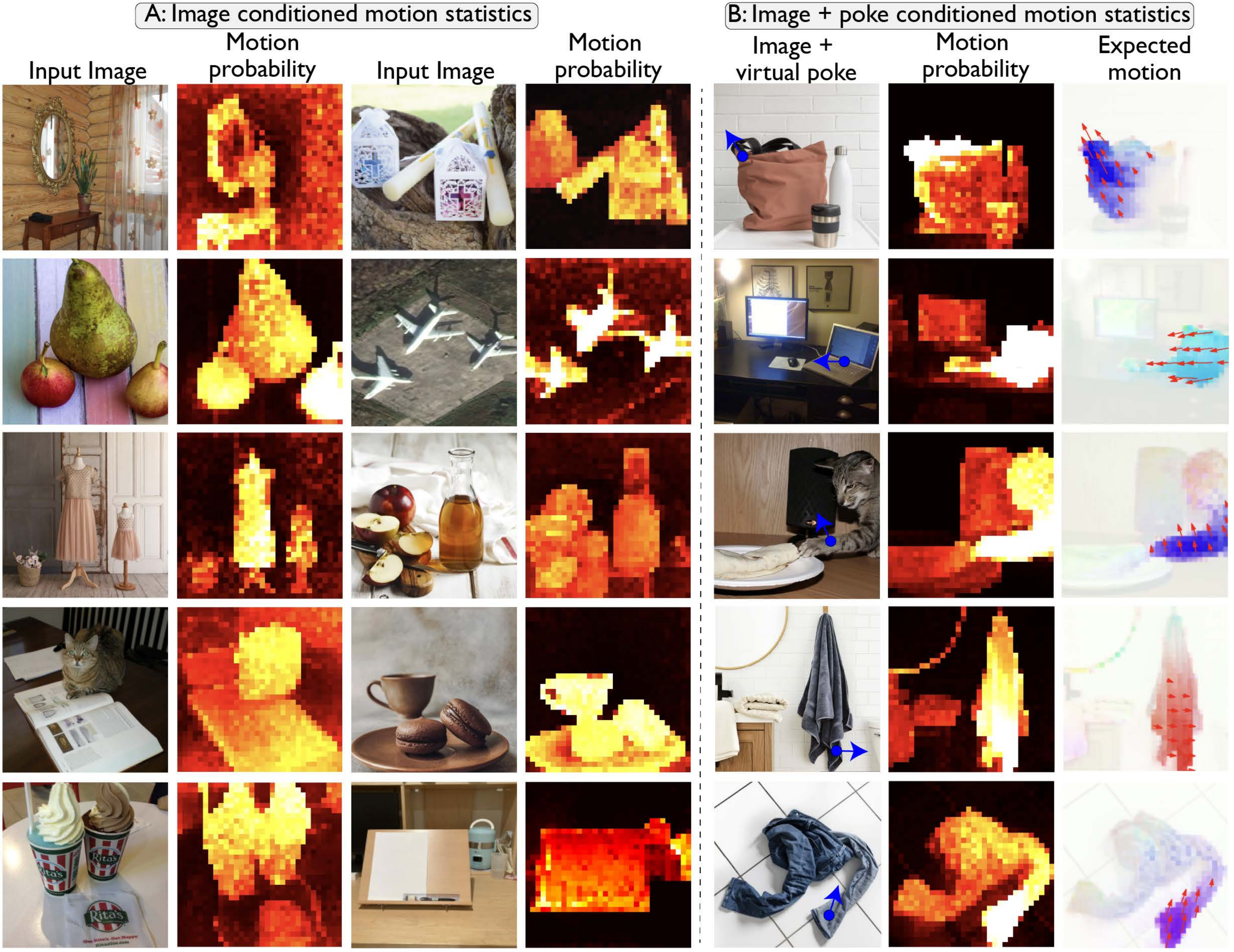}
  \caption{
  \textbf{Motion statistics computed in parallel.} In \textbf{Figure A}, we show motion probability computed given the input image and camera stop conditioning. They clearly highlight the parts of the scene that are likely to move. In \textbf{Figure B}, we show probability of motion and expected motion maps conditioned on an input virtual poke.} 
  \label{fig:supp_pmotion}
\end{figure*}

\begin{figure*}
  \centering
  \includegraphics[width=\linewidth]{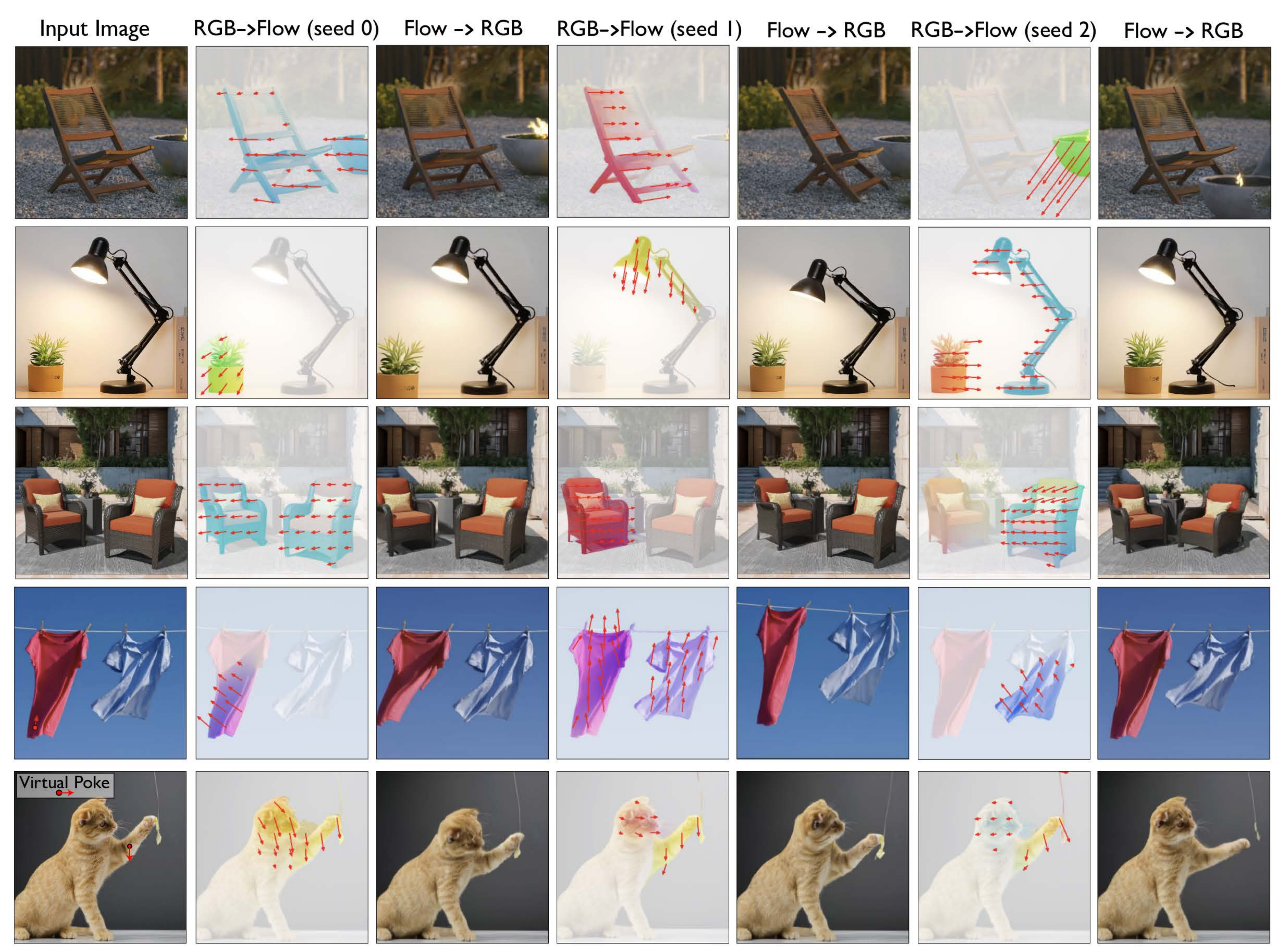}
  \vspace{-2em}
  \caption{
  \textbf{Sequential generation of plausible object dynamics and appearance.} We show that our model can generate multiple physically plausible scene motions and render them into future appearance states -- capturing the true dynamics of the physical world for complex objects. In rows 1–4, the model infers plausible motion patterns directly from a single input image. In the last row, specifying a motion for a part of the object (such as the hand of the cat) generates diverse, yet physically consistent responses for the rest of the body.} 
  \label{fig:supp_uncond}
  \vspace{-1.0em}
\end{figure*}

\begin{figure*}[!t]
  \centering
  \includegraphics[width=0.95\linewidth]{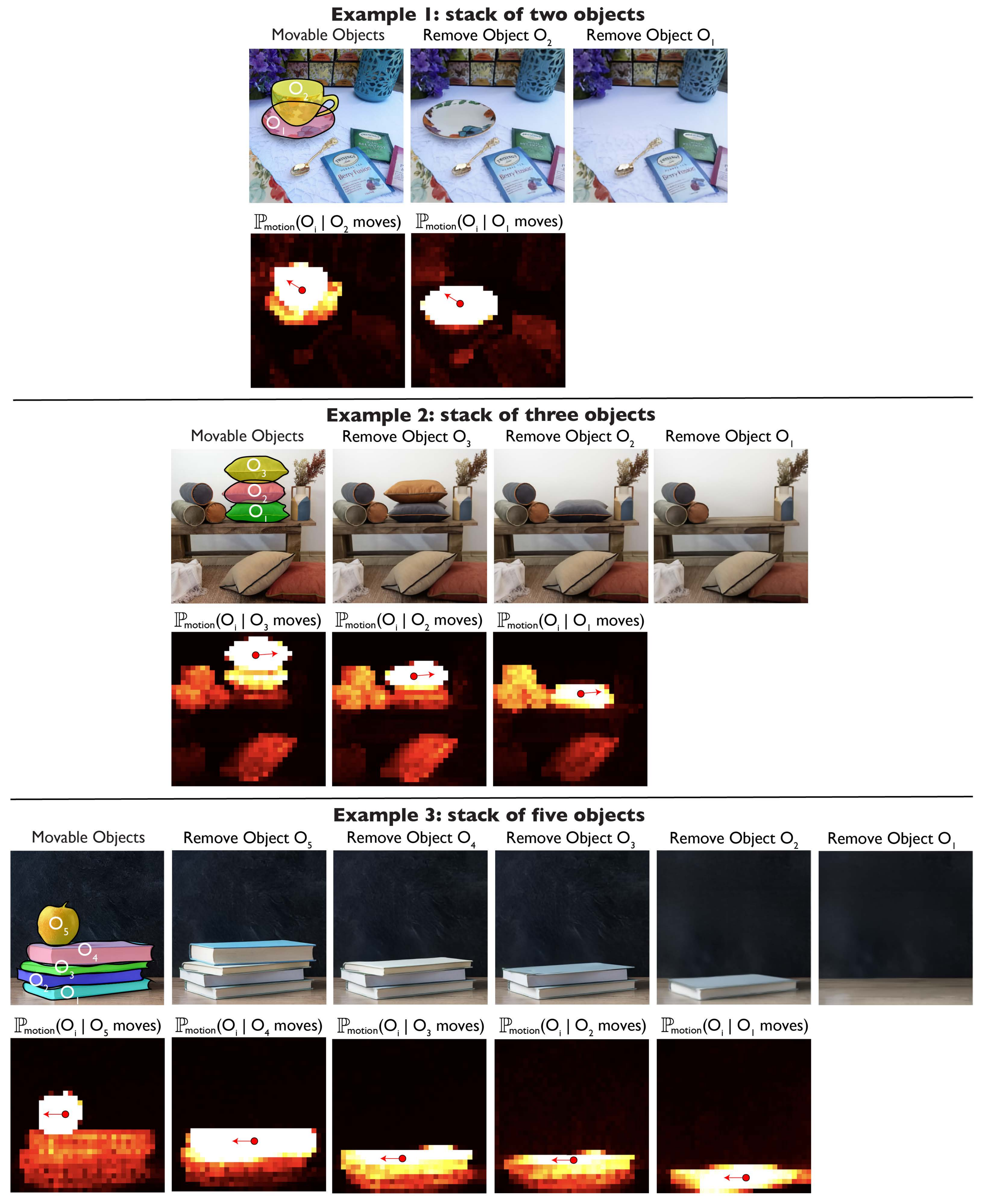}
  \caption{
  \textbf{Reasoning about physical relationships between objects.} Here we demonstrate how probability of motion maps can be used to probe physical dependencies in scenes, enabling applications like visual jenga on three progressively challenging real world examples.} 
  \label{fig:support_relationships}
  \vspace{-1.2em}
\end{figure*}

\begin{figure*}[htbp]  
  \centering
  \includegraphics[width=0.9\textwidth]{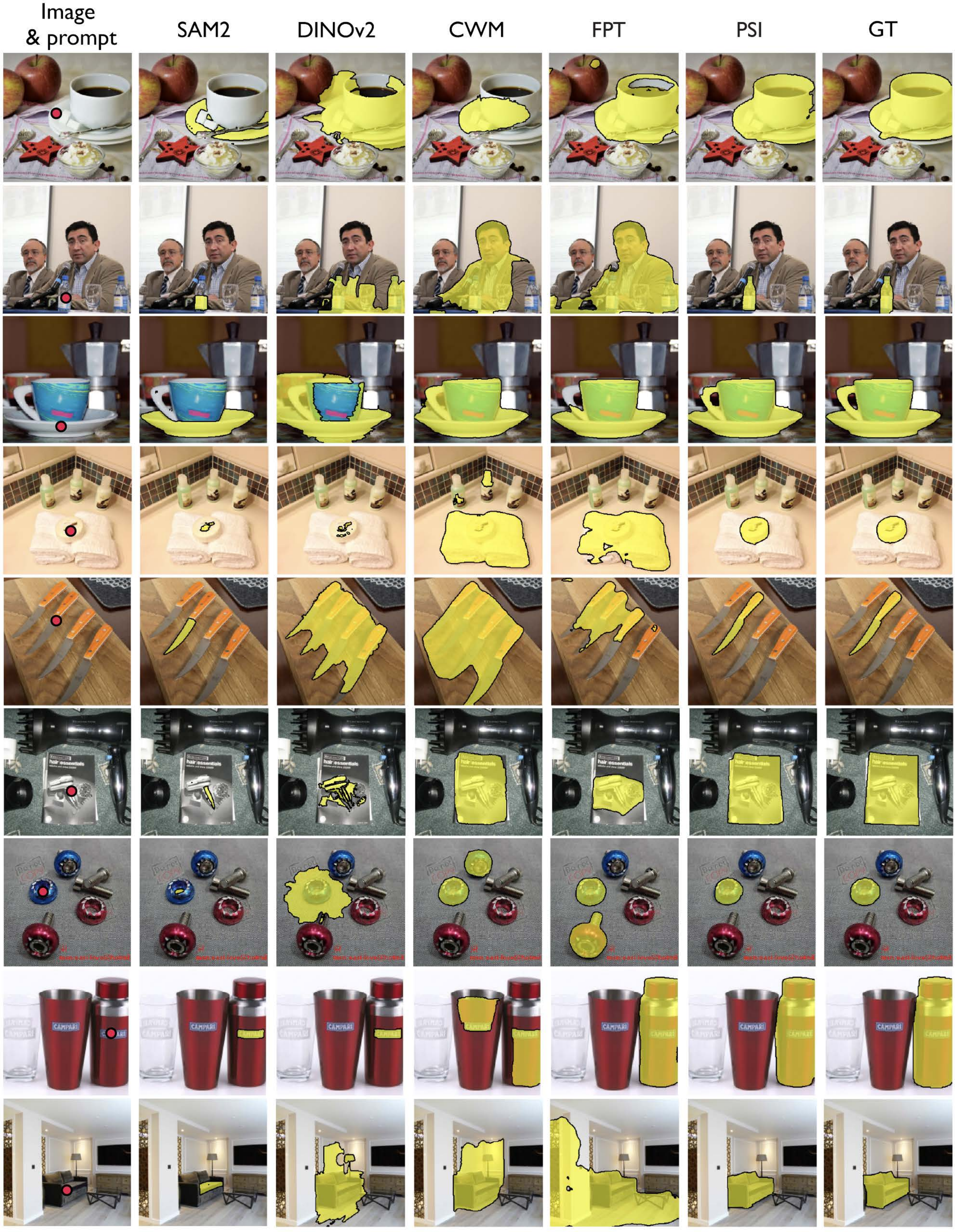}
  \vspace{-0.5em}
  \caption{\textbf{Additional qualitative results for point-promoted segmentation across models.} \lrasseg yields sharper segments which are more aligned with the notion of an object as a movable entity as compared to SAM2, DINO, CWM and FPT.}
  \label{fig:point_seg_supplementary}
  \vspace{-1.0em}
\end{figure*} 

\begin{figure*}[b!]
  \centering
  \includegraphics[width=\textwidth]{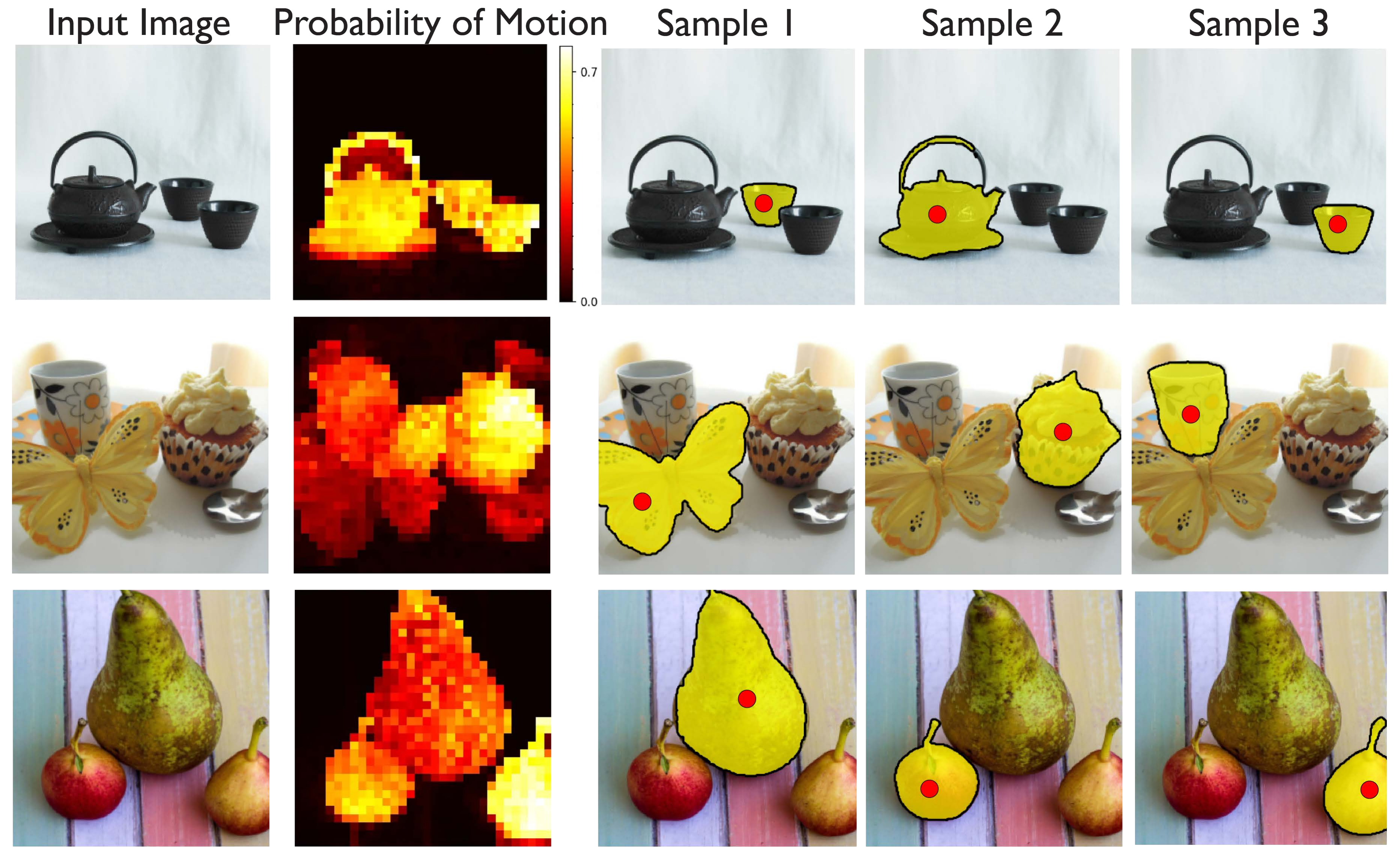}
  \caption{\textbf{Illustration of unprompted movable object segment discovery using \lrasseg.} The corresponding discovered segments are highlighted, demonstrating the ability of \lrasseg to automatically identify every movable object in the scene without manual prompts.}
  \label{fig:autoseg_suppl}
  \vspace{-1.0em}
\end{figure*}

\begin{figure*}[htbp]  
  \centering
  \includegraphics[width=0.9\textwidth]{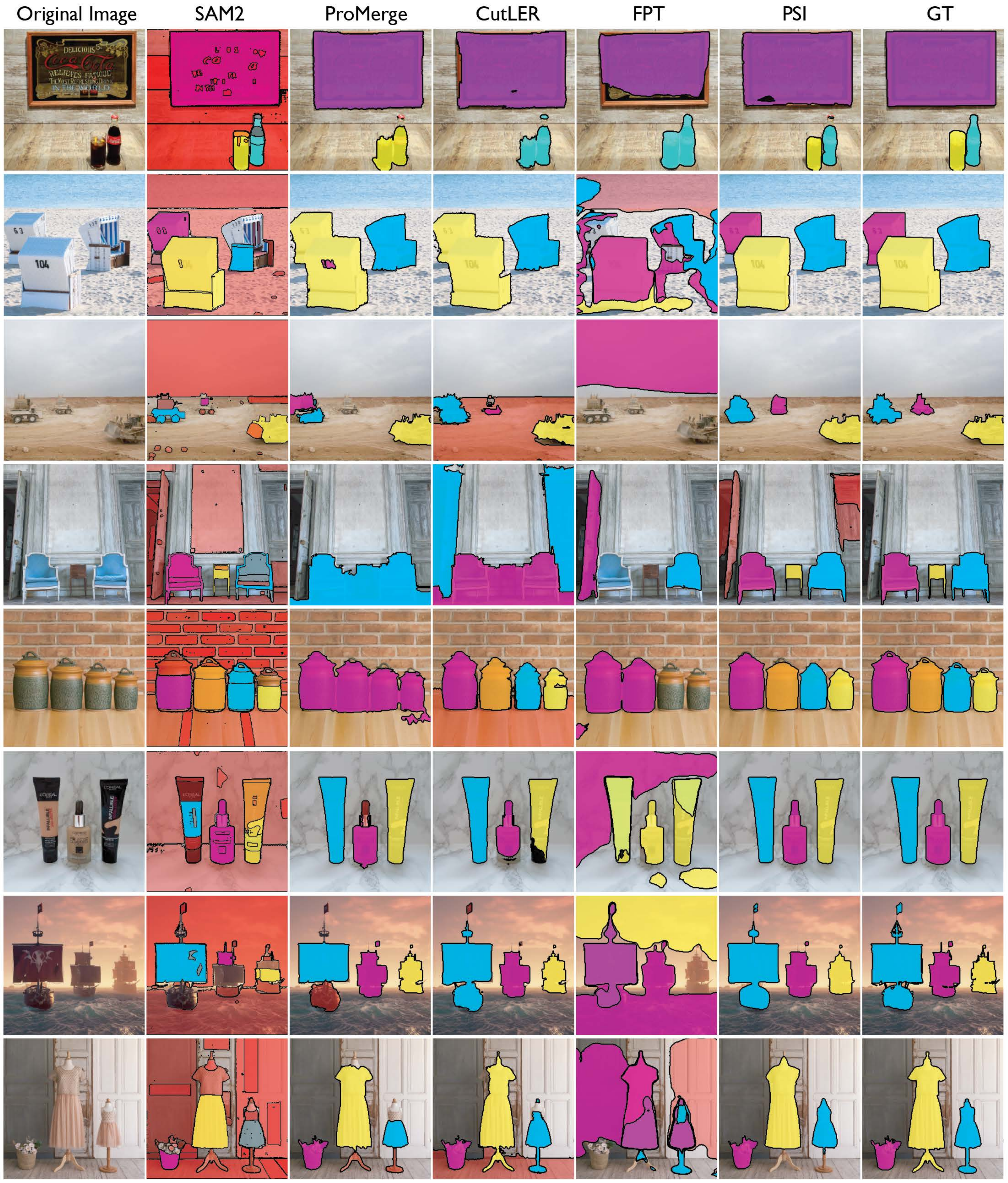}
  \vspace{-0.5em}
  \caption{\textbf{Additional qualitative results for automatic segmentation across models.} \lrasseg produces a set of physical object segments more consistent with physical co-movement as compared to SAM2, ProMerge, CutLER and FPT. Red regions denote predicted segments that are not matched to GT labels.}
  \label{fig:auto_seg_supplementary}
  \vspace{-1.0em}
\end{figure*} 

\begin{figure*}[htbp]  
  \centering
  \includegraphics[width=0.9\textwidth]{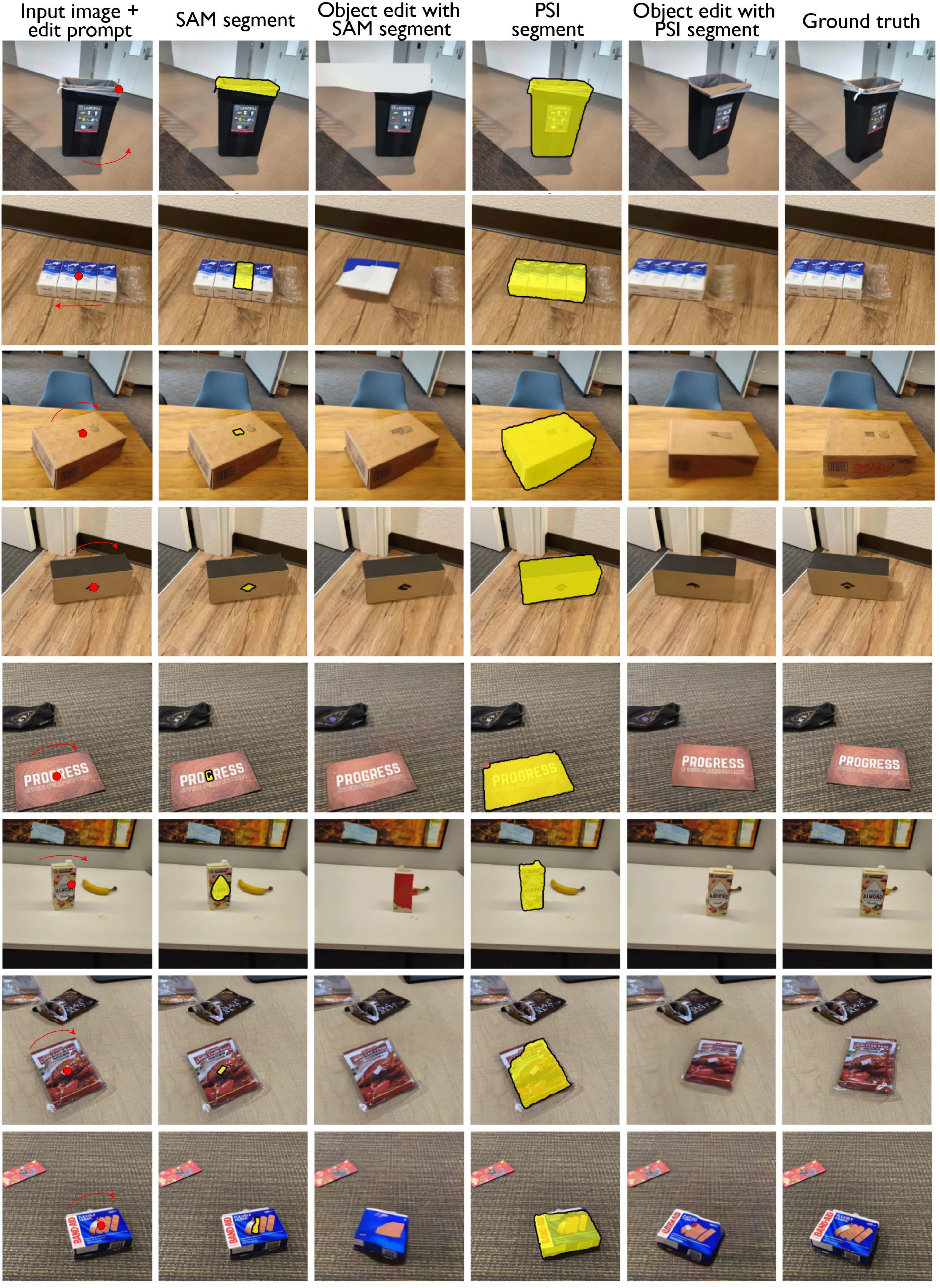}
  \vspace{-0.5em}
  \caption{\textbf{Additional qualitative comparisons of scene edits using SAM masks versus \lrasseg segments.} Each row shows the original image, the user click location, and the resulting edited image using different segmentation methods. }
  \label{fig:suppl_obj_manipulation}
  \vspace{-1.0em}
\end{figure*} 

\begin{figure*}[htbp]  
  \centering
  \includegraphics[width=0.9\textwidth]{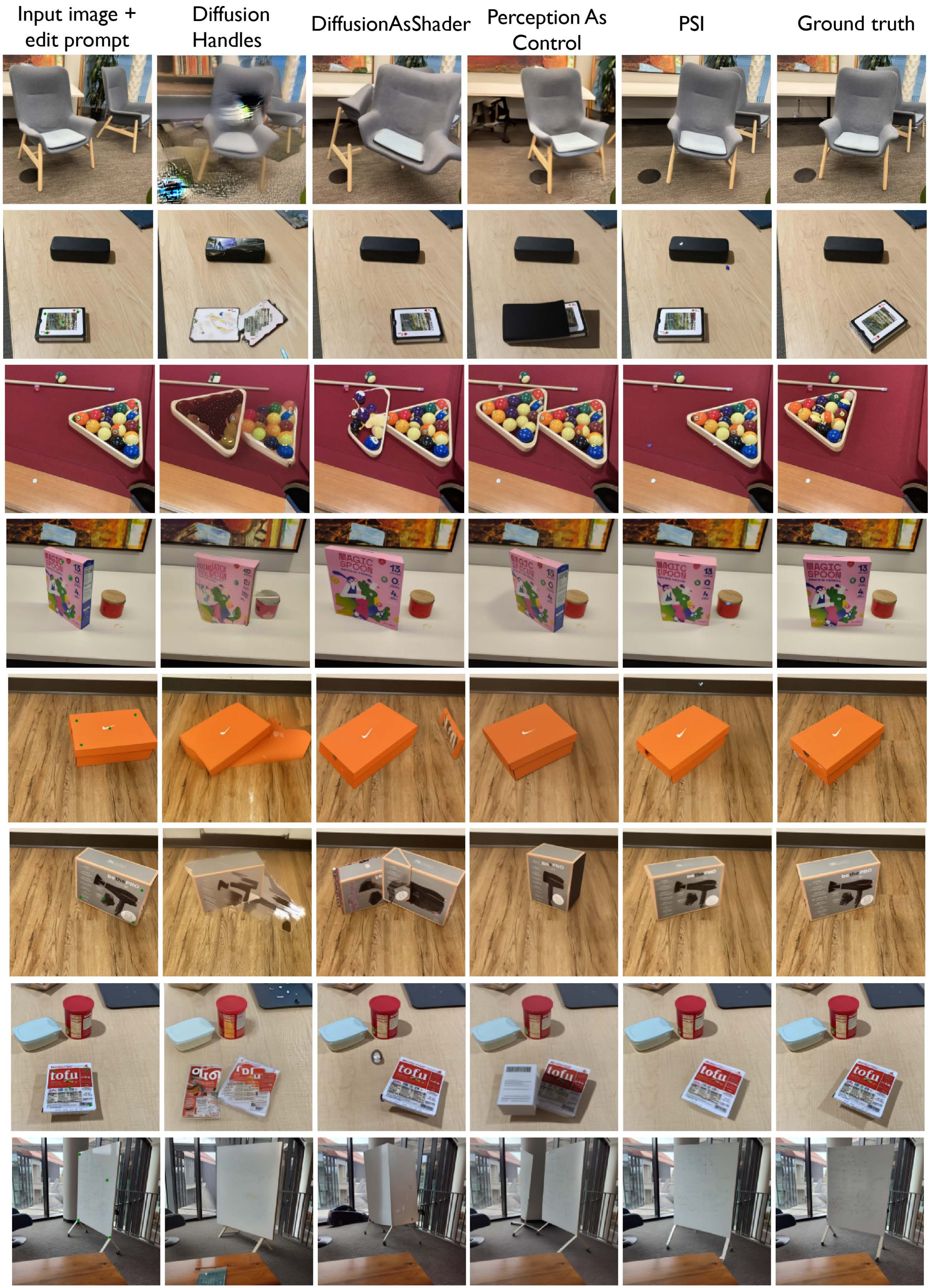}
  \vspace{-0.5em}
  \caption{\textbf{Additional qualitative comparisons of object manipulation comparisons with SOTA methods.} We compare \lrasseg to various object-centric image editing methods and show that \lrasseg enables more physically plausible image edits. }
  \label{fig:suppl_obj_manipulation_comp}
  \vspace{-1.0em}
\end{figure*} 

\begin{figure*}[htbp]  
  \centering
  \includegraphics[width=0.7\textwidth]{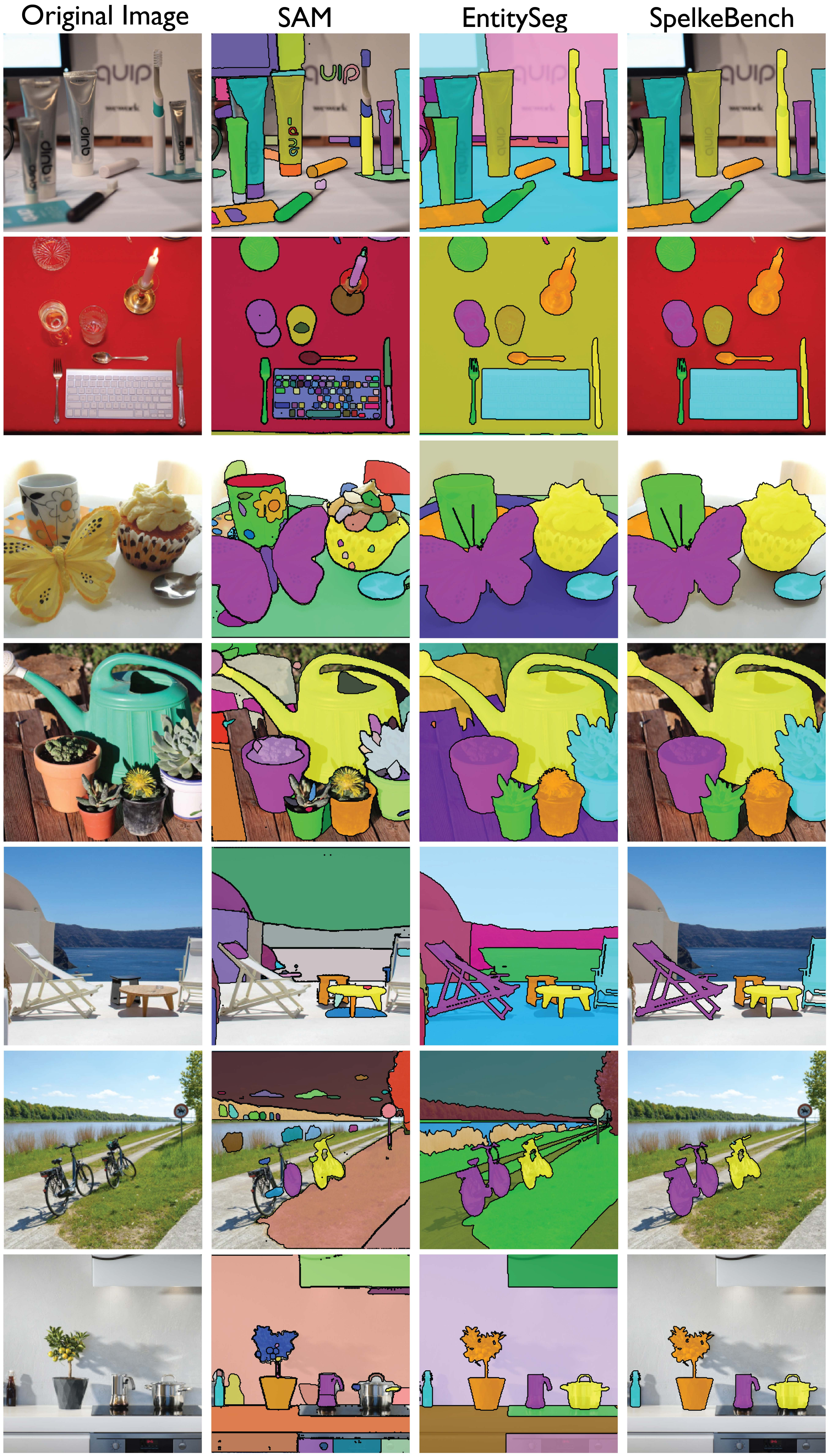}
  \vspace{-0.5em}
  \caption{\textbf{Qualitative comparison of \spelkeentity vs other datasets.} The visualization demonstrates characteristic differences across datasets: SAM's dataset tends to oversegment objects into constituent parts (i.e., bottle labels, cup designs), EntitySeg frequently includes ill-defined background regions (i.e., ground, wall), whereas \spelkeentity contains segments that better align with the notion of movable objects as units that move together, serving as an appropriate benchmark for the object discovery applications of our world model.}
  \label{fig:entityseg_dataset_suppl}
  \vspace{-1.0em}
\end{figure*}

\begin{figure*}[htbp]  
  \centering
  \includegraphics[width=0.87\textwidth]{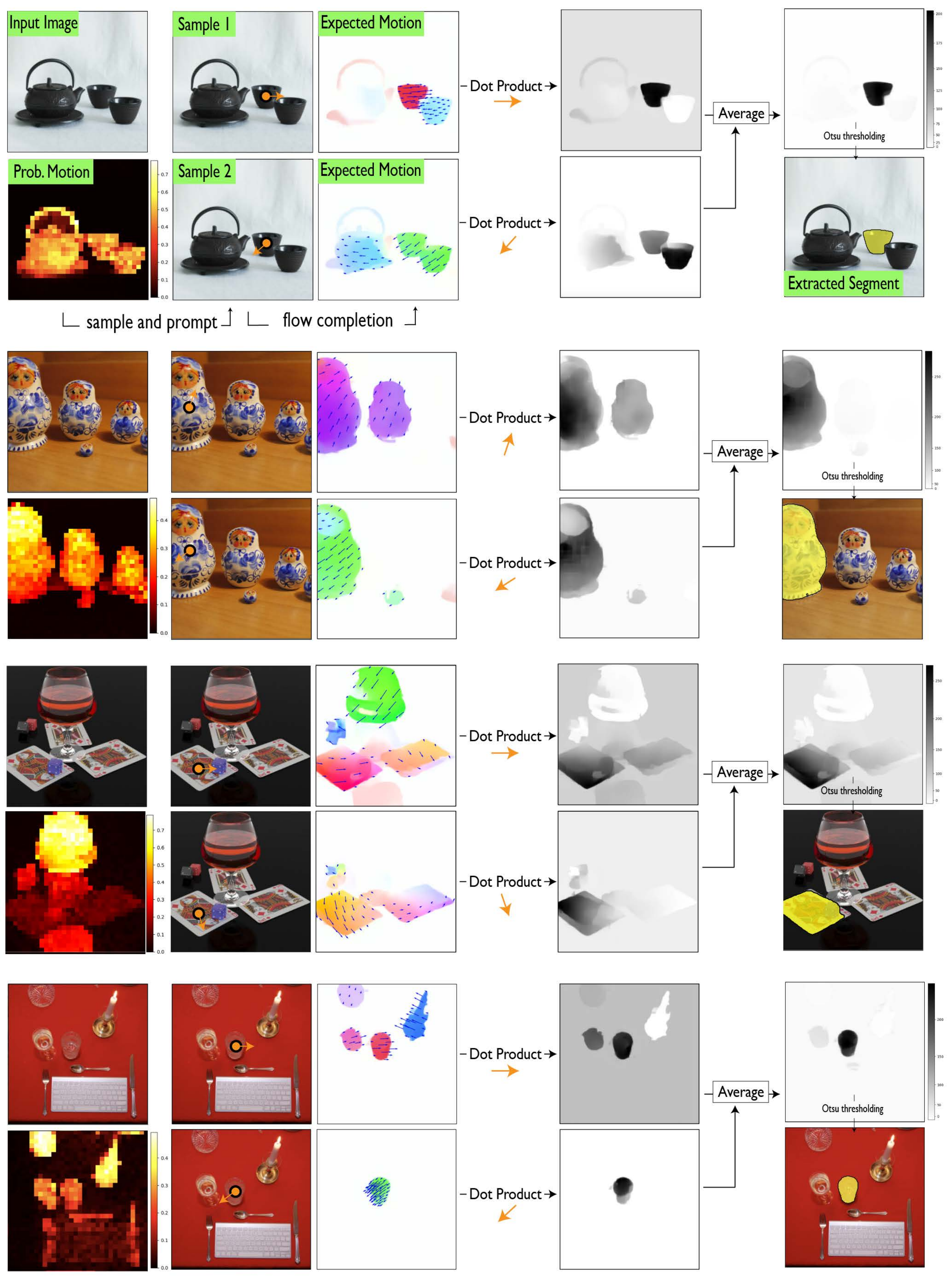}
  \vspace{-0.5em}
  \caption{\textbf{More illustrations of our movable object discovery algorithm described in Section 4.3 of the main paper}. To discover movable objects, we apply multiple virtual pokes at locations sampled from the $\mathbb{P}_{\text{motion}}$ map (column 2). While the model consistently propagates flow across the poked object (column 3), it also generates unprompted flow on other objects. However, since this unprompted flow varies across pokes and typically diverges in direction from the input poke, it gets suppressed when averaging the dot product (column 4) and helps us isolate independently movable entities as shown in the last column. Note that we average across 8 pokes, but only show two rows here for brevity.}
  \label{fig:point_seg_process}
  \vspace{-1.0em}
\end{figure*}

\begin{figure*}[b]
  \centering
  \includegraphics[width=\textwidth]{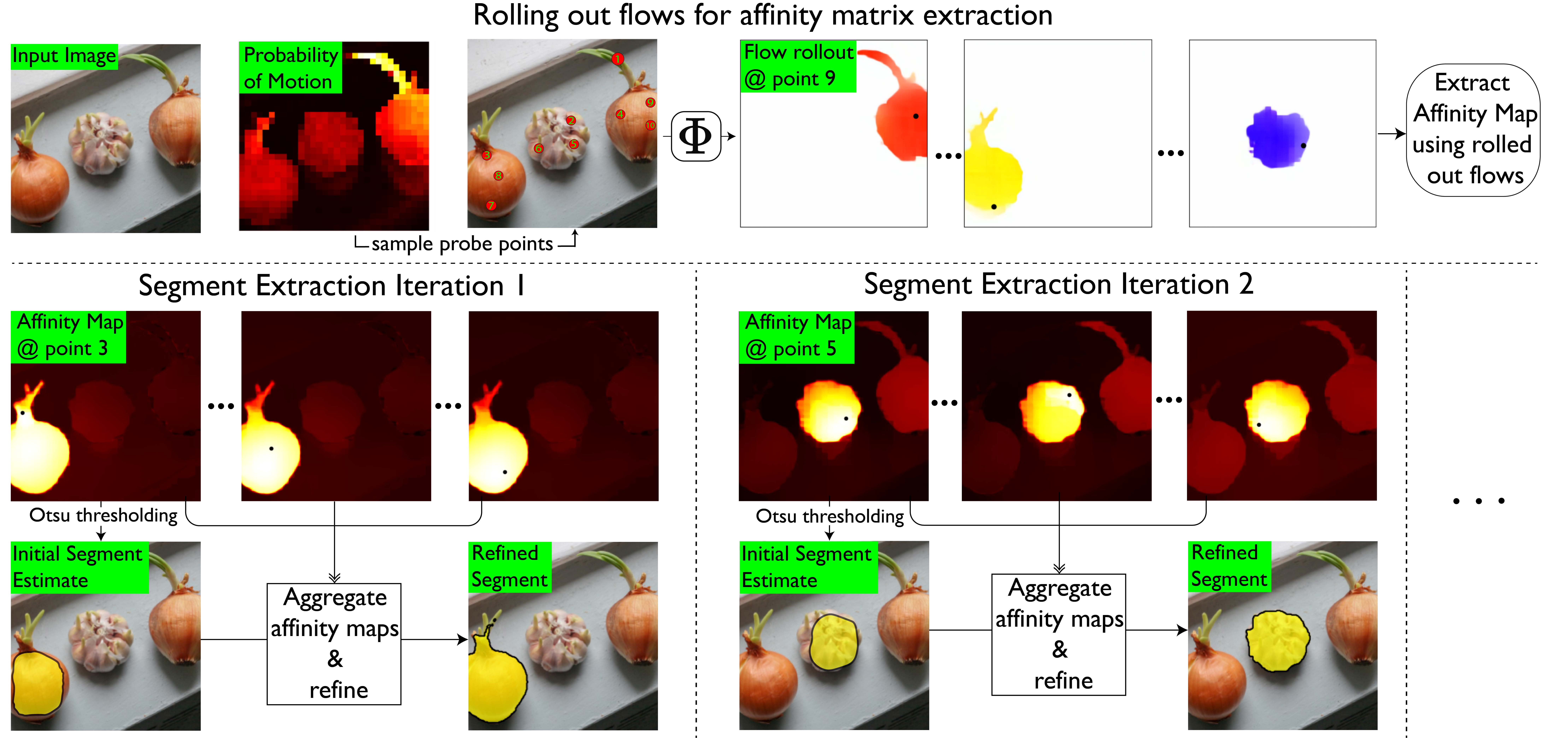}
  \vspace{-1em}
  \caption{\textbf{Unprompted discovery of movable object segments}. We extract probability of motion maps from an image, and use it to sample candidate poke points \textbf{(top left)}. We apply a poke to the image at the sampled points and obtain dense flow fields conditioned on the poke \textbf{(top right)} which are used to compute affinity maps. As shown in the \textbf{bottom} panel, these maps enable the extraction of segments using iterative clustering (see Section~\ref{sec:autoseg_procedure}).}
  \label{fig:autosegschematic}
  \vspace{-1.0em}
\end{figure*}


\begin{figure*}[b!]
  \centering
  \includegraphics[width=0.98\textwidth]{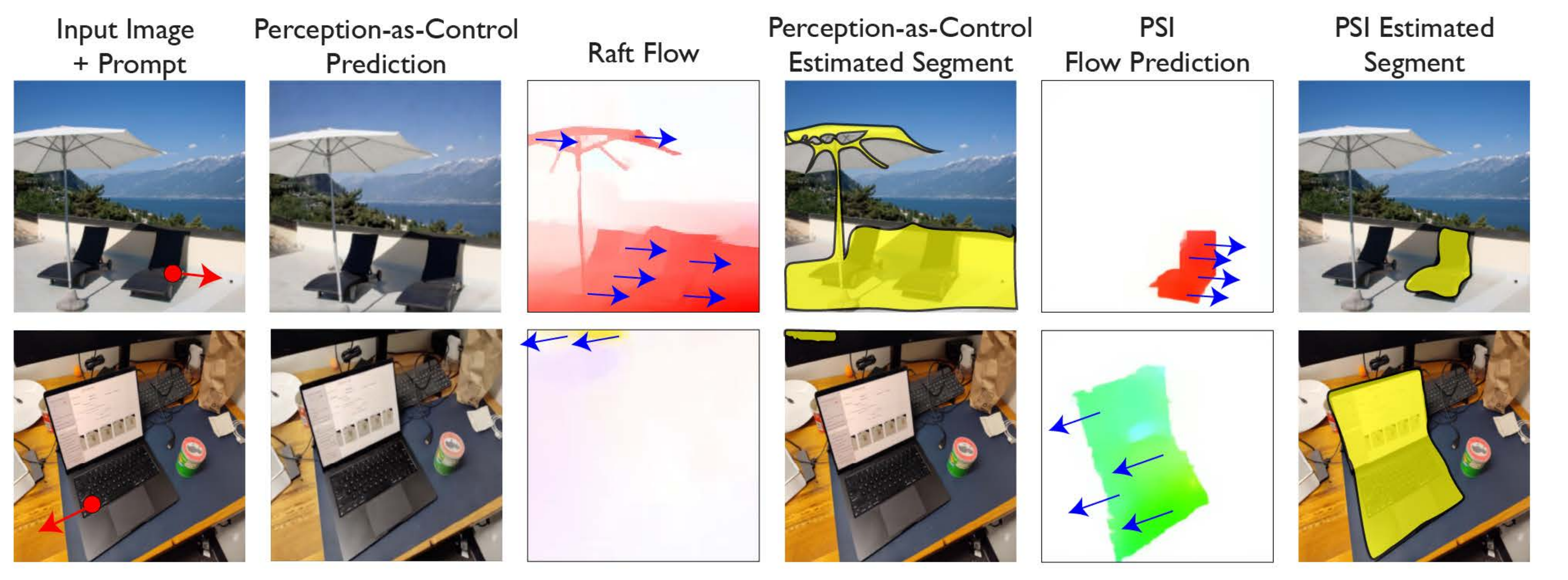}
  \vspace{1.0em}
  \caption{\textbf{Perception-as-Control~\cite{chen2025perception} failure modes in complex scenes.} The first column shows the input image with prompt, the second column displays the motion-controlled prediction from the Perception-as-Control, and the third column shows the RAFT-predicted flow field between the input and predicted image. The final column presents the resulting segment obtained by thresholding the flow magnitude. Compared to \lrasseg, Perception-as-Control often produces amorphous flow fields due to imprecise predictions based on the sparse motion control signals.}
  \label{fig:perceptionascontrol}
\end{figure*}

\begin{figure*}[b!]
  \centering
  \includegraphics[width=0.98\textwidth]{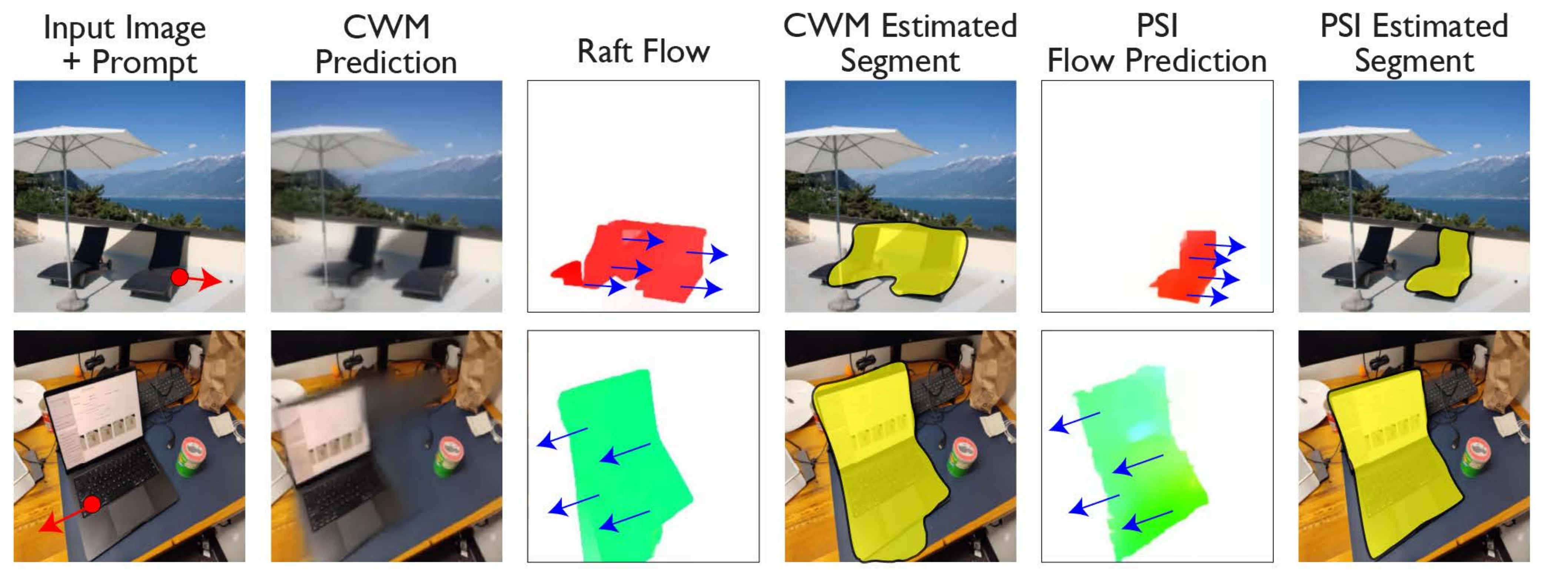}
  \vspace{1.0em}
  \caption{\textbf{CWM segmentation failure modes in complex scenes.} Each row shows a challenging example where CWM struggles. The first column shows the input image with the patch motion prompt (red arrow). The second column displays the counterfactual prediction generated by CWM. The third column shows the RAFT-predicted flow field between the input and counterfactual image. The final column presents the resulting segment obtained by thresholding the flow magnitude. Compared to \lrasseg, CWM often produces diffuse motion fields due to blurry RGB reconstruction and inaccurate object boundaries.}
  \label{fig:cwmcomparisonsuppl}
\end{figure*}


\clearpage
\clearpage
{
    \small
    \bibliographystyle{ieeenat_fullname}
    \bibliography{main}
}

\section*{Acknowledgements}
This work was supported by the following awards: To D.L.K.Y.: Simons Foundation grant 543061, National Science Foundation CAREER grant 1844724, National Science Foundation Grant NCS-FR 2123963, Office of Naval Research grant S5122, ONR MURI00010802, ONR MURI S5847, and ONR MURI 1141386- 493027. We also thank the Stanford HAI, Stanford Data Sciences and the Marlowe team, and the Google TPU Research Cloud team for computing support.

{
    \small
    \bibliographystyle{ieeenat_fullname}
    \bibliography{main}
}

\end{document}